\documentclass[runningheads]{llncs}
 
\usepackage{eccv}

\usepackage{eccvabbrv}

\usepackage{graphicx}
\usepackage{booktabs}
\usepackage[table,xcdraw]{xcolor}
\usepackage{multirow}
\usepackage{wrapfig}
\usepackage{CJK} 
\usepackage[linesnumbered,ruled,vlined]{algorithm2e}
\usepackage[accsupp]{axessibility}  

\definecolor{lightblue}{rgb}{0.9, 0.95, 1.0}
\definecolor{lightgray}{rgb}{0.9, 0.9, 0.9}

\usepackage[pagebackref,breaklinks,colorlinks,citecolor=eccvblue]{hyperref}

\usepackage{orcidlink}

\begin{document}
{
\renewcommand{\thefootnote}{}
\footnotetext{%
  \hspace{-1.8em}%
  \textdagger\ Corresponding author. Link: \href{https://github.com/wjl2244/BeyondDrive}{https://github.com/wjl2244/BeyondDrive}
}
\title{Beyond Imitation: Learning Safe End-to-End Autonomous Driving from Hard Negatives} 

\titlerunning{ }

\author{
Junli Wang$^{1,2}$,
Zhihua Hua$^{3}$,
Xueyi Liu$^{1}$,  
Zebin Xing$^{1}$, 
Haochen Tian$^{1,2}$,\\
Kun Ma$^2$,
Hangjun Ye$^2$,
Guang Chen$^{2}$, 
Long Chen$^{2}$,
Qichao Zhang$^{1 \dagger}$ \\
[3mm]
$^1$~Institute of Automation, Chinese Academy of Sciences \\
$^2$~Xiaomi Embodied Intelligence Team \quad
$^3$~Fudan University \\
[3mm]
}

\authorrunning{ }

\institute{
\begin{CJK}{UTF8}{gbsn}
\begin{quote}
见贤思齐焉，见不贤而内自省也。
\vspace{0.2em}
\raggedleft \textemdash 《论语·里仁》
\end{quote}
\end{CJK}
}

\maketitle
\vspace{-0.5cm}
\begin{figure}[!h]  
  \centering
  \includegraphics[height=5.3cm]{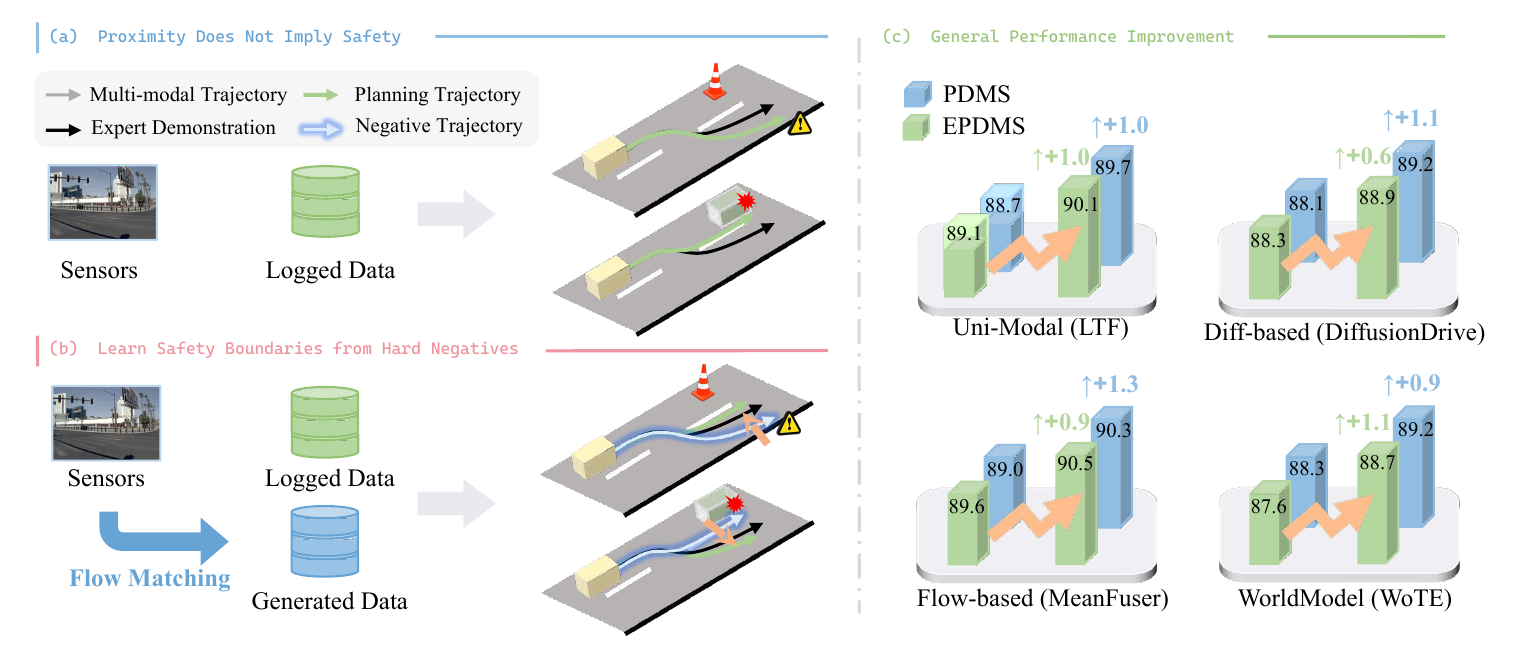}
  \caption{Learning safe autonomous driving from hard negatives. (a) Uni-modal and multi-modal end-to-end driving models learn by imitating expert demonstrations, yet they lack an understanding of "what constitutes poor imitation." (b) We generate unsafe hard negatives close to expert via flow matching, guiding models to avoid such behaviors. (c) Our approach yields significant performance improvements on reactive (EPDMS) and non-reactive (PDMS) closed-loop benchmarks for multiple methods.}
  \label{fig:teaser}
  \vspace{-0.8cm}
\end{figure}

\begin{abstract}

    Existing imitation learning methods for end-to-end autonomous driving predominantly learn from successful demonstrations by minimizing geometric deviations from expert trajectories. This paradigm implicitly assumes that spatial proximity implies behavioral safety, leading to a critical objective mismatch: trajectories with nearly identical imitation losses may exhibit drastically different safety outcomes, where one remains recoverable while the other results in collision. 
    To address this limitation, we propose \textbf{BeyondDrive}, a failure-aware imitation learning framework that jointly learns from successful and failed driving behaviors. First, we introduce a flow matching-based negative trajectory generator that synthesizes safety-critical yet expert-proximate trajectories, enabling explicit modeling of safety asymmetry. Second, we develop a diversity-aware sampling strategy that mitigates mode collapse and improves coverage of diverse failure modes during negative trajectory generation. Third, we propose a Repulsive Distance Loss that simultaneously attracts predictions toward expert demonstrations while repelling them from hard negative trajectories, thereby establishing discriminative safety boundaries in trajectory space. Applied to the uni-modal baseline Latent TransFuser, BeyondDrive achieves 89.7 PDMS on the NAVSIMv1 closed-loop benchmark, outperforming prior state-of-the-art methods. Moreover, BeyondDrive generalizes effectively across different autonomous driving architectures, including multi-modal planners, and further demonstrates strong zero-shot transferability on the HUGSIM benchmark.
    
\end{abstract}

\section{Introduction}

End-to-end autonomous driving is commonly formulated as trajectory imitation, where models are trained to minimize geometric deviations from expert demonstrations. Recent years have witnessed remarkable progress in this paradigm, with methods such as TransFuser~\cite{transfuser}, UniAD~\cite{uniad}, and VAD~\cite{vad} achieving strong planning performance through direct trajectory regression. However, these approaches rely on a fundamental assumption: trajectories that are geometrically close to expert demonstrations are also behaviorally safe. 

In practice, this assumption does not always hold. Driving exhibits a critical property that we term \textit{safety asymmetry}: trajectories with nearly identical imitation errors may lead to drastically different safety outcomes. As illustrated in \cref{fig:teaser}, two trajectories that deviate left or right from an expert demonstration can incur similar geometric losses, while one remains recoverable and the other results in collision. Since conventional imitation learning only optimizes toward successful demonstrations without explicitly modeling unsafe behaviors, the learned policy lacks clear safety boundaries between good and bad imitations.

To alleviate this issue, recent methods have shifted toward multi-modal trajectory generation and trajectory scoring. Methods such as VADv2~\cite{vadv2}, DiffusionDrive~\cite{diffusiondrive}, MeanFuser~\cite{meanfuser}, and WoTE~\cite{wote} generate multiple candidate trajectories using anchor-based designs or generative models~\cite{ddim,flowmatching}, followed by rule-based or learned scoring mechanisms for trajectory selection. Although these methods can identify unsafe or low-quality trajectories during inference, the identified failures are ultimately discarded after scoring and never explicitly participate in policy optimization. As a result, existing methods can recognize poor trajectories, but cannot effectively learn from them.

This naturally raises an important question: \textit{Can end-to-end driving policies explicitly learn from expert-proximate failures to establish discriminative safety boundaries?}

Contrastive learning~\cite{contrastive_learning} provides a natural perspective for addressing this problem, as its core principle is to simultaneously attract positive samples while repelling negative ones. However, constructing effective negative samples for autonomous driving is highly non-trivial. Useful negative trajectories must satisfy two conflicting properties: they should remain spatially close to expert demonstrations to form meaningful contrasts, while simultaneously containing explicit safety-critical behaviors that provide informative supervision. We refer to such trajectories as \textit{hard negatives}. Generating high-quality hard negatives therefore becomes the key challenge.

Based on these insights, we propose \textbf{BeyondDrive}, a failure-aware contrastive learning framework for end-to-end autonomous driving. BeyondDrive enables driving policies to explicitly learn safety boundaries by jointly leveraging successful demonstrations and synthesized hard negatives. Specifically, we first introduce a flow matching-based failure generator that actively synthesizes safety-critical trajectories near the expert manifold. To improve failure diversity and avoid mode collapse, we further incorporate diversity-aware sampling strategies during trajectory generation. Subsequently, we design a two-stage selection mechanism to identify informative negative trajectories that are both unsafe and geometrically close to expert demonstrations. Finally, we propose a Repulsive Distance Loss (RD Loss) that simultaneously pulls predictions toward expert trajectories while pushing them away from synthesized failures, enabling explicit contrastive safety optimization in trajectory space.

We apply BeyondDrive to Latent TransFuser (LTF)~\cite{navsim}, a uni-modal baseline on the NAVSIM benchmark~\cite{navsim}. BeyondDrive improves the baseline to 89.7 PDMS on NAVSIM v1, outperforming numerous recent autonomous driving methods. Furthermore, BeyondDrive further demonstrates strong generalizability across diverse multi-modal planning paradigms, including diffusion-based methods such as DiffusionDrive~\cite{diffusiondrive}, flow-based methods such as MeanFuser~\cite{meanfuser}, and world-model-based approaches such as WoTE~\cite{wote}, consistently yielding significant performance improvements. Additional zero-shot experiments on HUGSIM further demonstrate the robustness and generalizability of our framework.

The contributions of this work are summarized as follows:
\begin{itemize}
    \item We propose \textbf{BeyondDrive}, a failure-aware contrastive learning framework that explicitly learns safety boundaries using synthesized hard negatives.
    
    \item We introduce a flow matching-based failure generation strategy together with diversity-aware sampling and contrastive safety optimization, enabling effective construction and utilization of informative negative trajectories.
    
    \item Extensive experiments on NAVSIM and HUGSIM demonstrate that BeyondDrive consistently improves both uni-modal and multi-modal autonomous driving architectures, achieving strong closed-loop planning performance and robust generalization.
\end{itemize}

\section{Related Works}

In this section, we review end-to-end autonomous driving, the application of generative models in end-to-end autonomous driving, and the use of negative sample learning in end-to-end autonomous driving.
 
\subsection{End-to-End Autonomous Driving}

End-to-end autonomous driving~\cite{transfuser,uniad,diffusiondrive,consistencydrive,reasonplan,lu2026onevl,perlad,takevla} has evolved rapidly in recent years. Early methods such as TransFuser~\cite{transfuser} introduced multi-modal sensor fusion with direct trajectory regression for interpretable planning, while subsequent frameworks including UniAD~\cite{uniad} and VAD~\cite{vad} further unified perception, prediction, and planning into differentiable end-to-end architectures. Despite their architectural differences, these methods predominantly follow an imitation-based learning paradigm, where policies are optimized by minimizing geometric deviations from expert demonstrations. As a result, supervision is restricted to successful driving behaviors, and unsafe trajectories are never explicitly modeled during training. To alleviate the limitations of direct regression, recent works have increasingly adopted candidate generation and trajectory scoring paradigms. DiffusionDrive~\cite{diffusiondrive} explores diffusion-based trajectory generation, MeanFuser~\cite{meanfuser} adopts flow-based generative planning, while VADv2~\cite{vadv2} and Hydra-MDP~\cite{hydra-mdp} generate trajectory candidates from predefined vocabularies and select optimal trajectories using learned scoring functions. WoTE~\cite{wote} further improves planning robustness through world-model-based trajectory evaluation and ensemble scoring strategies. Although these methods can identify low-quality or unsafe trajectories during inference, the corresponding failure signals are only used for trajectory ranking and selection. Unsafe trajectories identified by the scoring module are ultimately discarded and do not explicitly participate in policy optimization. Consequently, existing methods can recognize undesirable behaviors, but remain unable to directly learn safety boundaries from them.



\subsection{Negative Sample Learning in Autonomous Driving}

Negative sample learning has been widely adopted for representation learning and safety-critical understanding. For trajectory prediction, AMD~\cite{amd} and TrACT~\cite{tract} enhance rare scenario recognition via decoupled contrastive learning and prototypical clustering, while ECAM~\cite{ecam} and Drive-CLIP~\cite{drive-clip} incorporate negative signals for collision avoidance and risk assessment. SimScale~\cite{simscale} further studies simulation-based negative data generation but is computationally expensive. DriveDPO~\cite{drivedpo} and TakeAD~\cite{takead} extend this idea to preference-based and takeover-based learning, but they operate on pre-collected or predefined trajectory sets, where supervision is limited to ranking or selection over fixed candidates. As a result, unsafe yet expert-proximate trajectories are never explicitly incorporated into policy optimization, preventing the model from learning fine-grained safety boundaries in continuous trajectory space. BeyondDrive addresses this limitation by introducing such expert-proximate negative trajectories into training and enabling contrastive optimization over the trajectory generation process.

\section{Preliminary}

In this section, we first introduce flow matching and classifier-free guidance.

\subsection{Flow Matching.}

Flow Matching (FM) is a simulation-free technique for training Continuous Normalizing Flows (CNFs) by directly regressing vector fields \cite{flowmatching}. The core idea is to define a probability path connecting a simple prior distribution $p_0$ (e.g., standard Gaussian) and the target data distribution $p_1$, and learn a time-dependent velocity field $v_t: [0,1] \times \mathbb{R}^d \rightarrow \mathbb{R}^d$ that generates this path \cite{flowmatching, building_normalizing}. Given a coupling of noise samples $\mathbf{x}_0 \sim p_0$ and data samples $\mathbf{x}_1 \sim p_1$, a common choice is the conditional linear interpolation path: $\mathbf{x}_t = (1-t)\mathbf{x}_0 + t\mathbf{x}_1$, which yields a target velocity $\mathbf{u}_t = \mathbf{x}_1 - \mathbf{x}_0$ \cite{building_normalizing, rectifiedflow}. The FM objective minimizes the expected squared error between the predicted velocity $v_t(\mathbf{x}_t)$ and this target:
\begin{equation}
\mathcal{L}_{\mathrm{FM}} = \mathbb{E}_{t \sim \mathcal{U}(0,1), \mathbf{x}_0 \sim p_0, \mathbf{x}_1 \sim p_1} \left\| v_t(\mathbf{x}_t,t) - (\mathbf{x}_1 - \mathbf{x}_0) \right\|_2.
\end{equation}
where $\mathcal{U}(0,1)$ denotes the uniform distribution in the interval $[0,1]$. Once trained, samples are generated by solving the Ordinary Differential Equation (ODE) $\mathrm{d}\mathbf{x}_t = v_t(\mathbf{x}_t,t) \mathrm{d}t$ from $t=0$ to $t=1$ starting from random noise \cite{ODE, flowmatching}. Compared to diffusion models, FM offers faster sampling and simpler training objectives \cite{flowmatching, latentflow}.

\subsection{Classifier-Free Guidance.}

Classifier-Free Guidance (CFG) \cite{cfg} is a technique originally proposed for diffusion models to trade off diversity and fidelity in conditional generation. It jointly trains a conditional and an unconditional model by randomly dropping the condition during training, and extrapolates the score estimate during sampling. Recent works extend CFG to flow matching, forming Guided Flows \cite{meanflow, latentflow}. Analogous to diffusion CFG, the guided velocity field is computed as:
\begin{equation}
\tilde{v}_t(\mathbf{x}_t,t|\mathcal{C}) = v_t(\mathbf{x}_t,t) + w \cdot \big( v_t(\mathbf{x}_t,t|\mathcal{C}) - v_t(\mathbf{x}_t,t) \big),
\end{equation}
where $v_t(\mathbf{x}_t,t|\mathcal{C})$ and $v_t(\mathbf{x}_t,t)$ are conditional and unconditional velocity estimators, respectively. The guidance scale $w\geq0$ balances conditional fidelity and sample diversity: $w=1$ gives standard conditional generation, $w>1$ amplifies the condition for higher fidelity, and $w<1$ increases diversity. This flexibility makes CFG particularly suitable for generating diverse trajectory candidates that remain close to expert demonstrations, a key requirement for constructing informative negative samples in our framework.

\section{Method}

In this section, we briefly revisit the end-to-end autonomous driving task and introduce our proposed framework, \textbf{BeyondDrive}, which enhances the safety of imitation learning. As shown in \cref{fig:pipeline}, BeyondDrive comprises two key stages: first, we generate diverse unsafe yet expert-proximate negative samples using a flow matching-based generator with a diversity enhancement strategy that combines CFG and noise std scaling to prevent mode collapse; second, we employ a RD loss to pull predictions toward expert demonstrations while pushing them away from hard negative, encouraging better separation between safe and unsafe behaviors through contrastive learning.

\begin{figure}[t]  
  \centering
      \includegraphics[height=6.0cm]{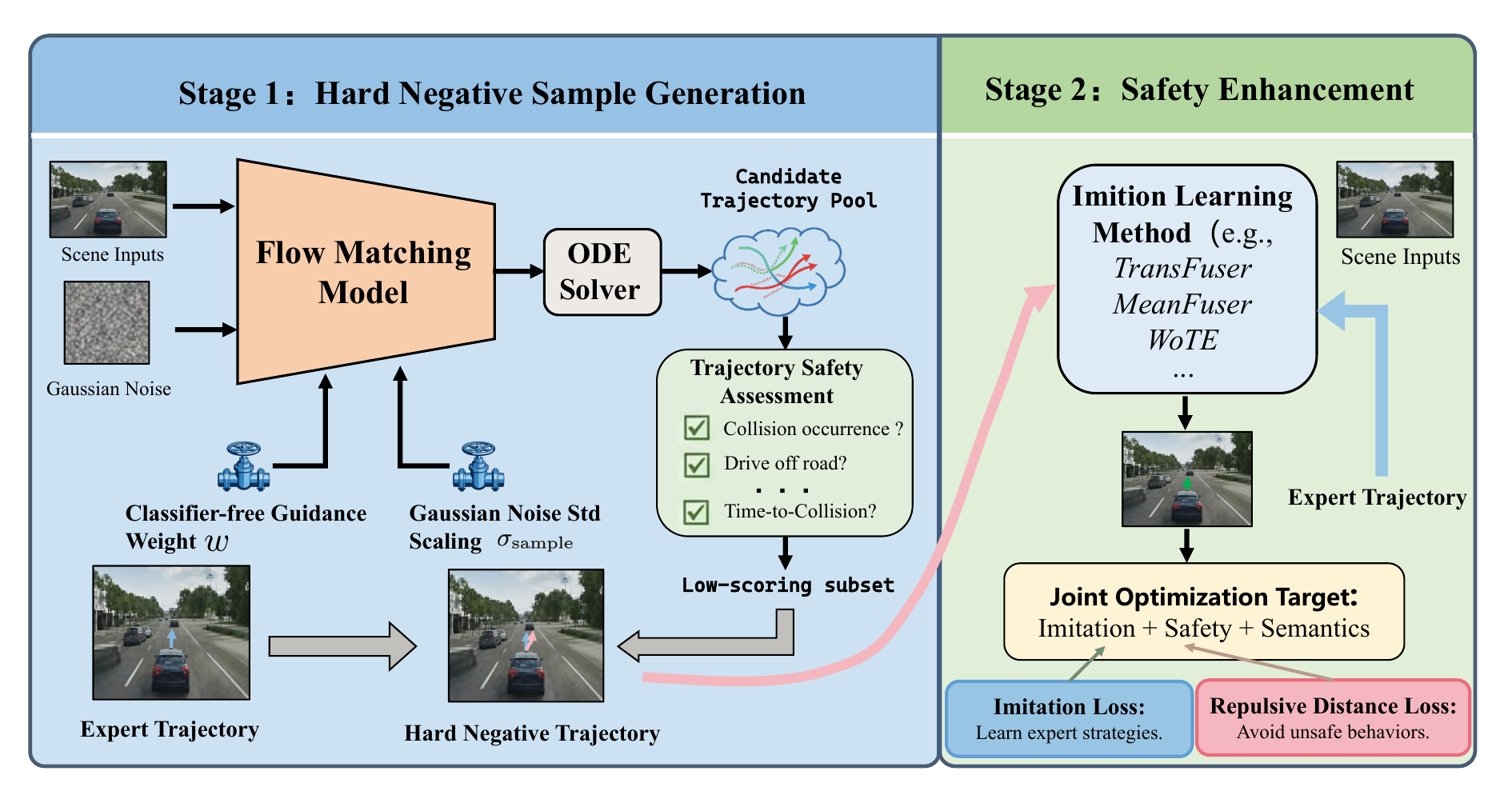}
  \caption{Overview of the BeyondDrive framework. In Stage 1, classifier-free guidance and noise standard deviation scaling are employed to generate diverse trajectory candidates near expert demonstrations, followed by safety-aware and distance-aware filtering to construct hard negative samples that are unsafe yet expert-proximate. In Stage 2, the imitation learning loss and repulsive distance loss jointly optimize the policy to learn expert strategies while avoiding unsafe behaviors.}
  \label{fig:pipeline}
\end{figure}

\subsection{Problem Formulation}

End-to-end autonomous driving systems take raw sensor data as input and directly plan a future trajectory. At each time step $t$, the ego vehicle observes multi-modal sensory data: a front-view RGB image $\mathbf{I}_t \in \mathbb{R}^{H \times W \times 3}$, a LiDAR point cloud $\mathbf{P}_t \in \mathbb{R}^{N \times 3}$ (with $N$ points), and ego-vehicle states $\mathbf{s}_t = [v_t, a_t]$ consisting of velocity and acceleration. Additionally, a high-level navigation command $c_t$ (e.g., go straight, turn left, turn right) is provided by a global planner. The model is required to output a planned trajectory ${{\hat \tau}_t} = \{ (x_{t+k\Delta t}, y_{t+k\Delta t}, h_{t+k\Delta t}) \}_{k=1}^{K}$ covering a $4$-second horizon at a $2$\,Hz control frequency, where $K=8$ and $\Delta t = 0.5$\,s. Each waypoint is represented in the ego-vehicle coordinate frame with position $(x,y)$ and heading angle $h$.

\subsection{Guided Trajectory Generation via Flow Matching}

To generate high-quality negative samples, we first train a conditional flow matching model for trajectory generation. Let $\mathbf{x}_0 \sim \mathcal{N}(\mathbf{0}, \sigma^2\mathbf{I})$ be a noise sample and let $\mathbf{x}_1 = \mathrm{vec}(\boldsymbol{\tau}_{\text{exp}})$ be the vectorized representation of an expert demonstration $\boldsymbol{\tau}_{\text{exp}}$ from the dataset. The context features $\mathcal{C} = \mathcal{E}_{\theta}(\mathbf{I}_t, s_t, c_t)$ are extracted by a scene encoder $\mathcal{E}_{\theta}$ \cite{transfuser} from the front-view image $\mathbf{I}_t$, ego states $s_t$, and navigation command $c_t$. We adopt the conditional linear interpolation $\mathbf{x}_t = (1-t)\mathbf{x}_0 + t\mathbf{x}_1$ with the corresponding target velocity $\mathbf{u}_t = \mathbf{x}_1 - \mathbf{x}_0$.

The model learns a time-dependent velocity field $v_\theta(\mathbf{x}_t,t|\mathcal{C}_{\text{in}})$ conditioned on context features. During training, we randomly drop the condition with probability $p=0.1$ to enable classifier-free guidance: $\mathcal{C}_{\text{in}} = \mathcal{C}$ with probability $1-p$, and $\mathcal{C}_{\text{in}} = \emptyset$ otherwise. We train the network $v_\theta$ with an $\ell_1$ loss instead of the conventional mean squared error:
\begin{equation}
\mathcal{L}_{\mathrm{FM}} = \mathbb{E}_{t\sim\mathcal{U}(0,1),\mathbf{x}_0,\mathbf{x}_1}\left\| v_\theta(\mathbf{x}_t,t|\mathcal{C}_{\text{in}}) - (\mathbf{x}_1 - \mathbf{x}_0) \right\|_1,
\end{equation}
which yields better generation fidelity in planning. The network architecture $v_\theta$ consists of a Transformer Decoder, LayerNorm, and a Feed-Forward Network.

To achieve broader coverage of the high-quality hard negative space, we control the standard deviation by setting the prior noise scale to $\sigma$ during training and increasing the sampling range to $\sigma_{sample} = 2.0\sigma$ during inference. At inference, we employ classifier-free guidance to steer the generation toward regions of high conditional likelihood. The guided velocity field is computed as:
\begin{equation}
\tilde{v}_\theta(\mathbf{x}_t,t|\mathcal{C}_{\text{in}}) = v_\theta(\mathbf{x}_t,t) + w \big( v_\theta(\mathbf{x}_t,t|\mathcal{C}_{\text{in}}) - v_\theta(\mathbf{x}_t,t) \big),
\end{equation}
where $v_\theta(\mathbf{x}_t,t)$ is the unconditional velocity estimate obtained by dropping the condition, and $w \geq 0$ is the guidance scale. For each conditioning input $\mathcal{C}_{\text{in}}$, we sample $64$ candidate trajectories $\{\boldsymbol{\tau}_i\}_{i=1}^{64}$ in parallel by solving the ODE $\mathrm{d}\mathbf{x}_t = \tilde{v}_\theta(\mathbf{x}_t,t|\mathcal{C}_{\text{in}})\mathrm{d}t$ from $t=0$ to $t=1$ and reshaping the resulting $\mathbf{x}_1$ back to trajectory form.
We then assess each candidate using a scoring function $S(\boldsymbol{\tau})$ (\eg, PDMS) that measures safety and driving quality. To construct informative negative samples, we first select a subset of trajectories with the lowest $S(\boldsymbol{\tau})$:
\begin{equation}
\mathcal{U}_{\text{low}} = \{\boldsymbol{\tau} \in \{\boldsymbol{\tau}_i\} \mid S(\boldsymbol{\tau}) < \xi \},
\end{equation}
where $\xi$ is a threshold that retains the most unsafe candidates. From this set, we pick the trajectory that is spatially closest to the expert demonstration $\boldsymbol{\tau}^{\text{exp}}$:
\begin{equation}
\boldsymbol{\tau}^{\text{neg}} = \mathop{\arg\min}_{\boldsymbol{\tau} \in \mathcal{U}_{\text{low}}} \left\| \boldsymbol{\tau} - \boldsymbol{\tau}^{\text{exp}} \right\|_2,
\end{equation}
where $\|\cdot\|_2$ denotes the Euclidean distance summed over all waypoints. This two-stage selection ensures that the negative samples are both unsafe and semantically similar to expert demonstrations, providing challenging contrasts for the subsequent policy learning stage. The training algorithm follows the procedure outlined in \cref{alg:training}, while the hard negative sampling process is described in \cref{alg:sample}.

\subsection{Negative Sample Learning}

We apply our BeyondDrive framework to the baseline TransFuser~\cite{transfuser} model, obtaining an enhanced version denoted as TransFuser v7. TransFuser takes as input the front-view image $\mathbf{I}_t$, LiDAR point cloud $\mathbf{P}_t$, ego state $\mathbf{s}_t$, and navigation command $c_t$. It employs a ResNet-34 backbone to extract context features $\mathcal{C}$ and is trained with an auxiliary map reconstruction loss $\mathcal{L}_{\text{map}}$\cite{navsim} on rasterized HD maps, which facilitates learning better scene representations.

To smooth the data distribution, we normalize using the first-order difference of trajectory coordinates, denoted as $\Delta \boldsymbol{\tau} = \{ (\Delta x_k, \Delta y_k, \sin h_k, \cos h_k) \}_{k=1}^{K}$. Here, $\Delta x_k$ and $\Delta y_k$ are the displacements over the $k$-th time interval (equivalent to average velocity scaled by $\Delta t = 0.5\,\text{s}$), while $\sin h_k$ and $\cos h_k$ encode the heading angle. This parameterization avoids the discontinuity inherent in direct angle regression, resulting in a smoother learning target that facilitates network convergence. The trajectory loss is then defined as the $\ell_1$ distance between the predicted and ground-truth differentials:
\begin{equation}
\mathcal{L}_{\text{exp}} = \| \Delta\hat{\boldsymbol{\tau}} - \Delta\boldsymbol{\tau}^{\text{exp}}\|_1. 
\end{equation}

To inject safety awareness, we introduce a RD loss that encourages the predicted trajectory to stay away from unsafe yet spatially similar negative samples generated by our flow matching module. Given a negative trajectory $\boldsymbol{\tau}_{\text{neg}} = \{(x_k^{\text{neg}}, y_k^{\text{neg}}, h_k^{\text{neg}})\}_{k=1}^{K}$ associated with the same condition, we compute its differential counterpart $\Delta\boldsymbol{\tau}^{\text{neg}}$ in the same manner. The RD loss is then defined as the negative clipped $\ell_1$ distance between the predicted and the hard negative differentials, where the distance is upper-bounded by a constant \(C\) :
\begin{equation}
\mathcal{L}_{\text{rd}} = -\min (\left\| \Delta\hat{\boldsymbol{\tau}} - \Delta\boldsymbol{\tau}^{\text{neg}} \right\|_1, C),
\end{equation}
this formulation pushes the prediction away from unsafe regions in the trajectory space, while the trajectory imitation loss $\mathcal{L}_{\text{imi}}$ simultaneously pulls it toward the expert demonstration, thereby improving the separability between safe and unsafe trajectories through contrastive optimization.
The overall training objective combines all components:
\begin{equation}
\mathcal{L} = \lambda_{\text{imi}}\mathcal{L}_{\text{imi}} + \lambda_{\text{rd}} \mathcal{L}_{\text{rd}} + \lambda_{\text{sem}} \mathcal{L}_{\text{sem}}  ,
\end{equation}
where $\lambda_{\text{imi}}$, $\lambda_{\text{rd}}$, and $\lambda_{\text{sem}}$ are weighting coefficients balancing the contributions of each term.

\section{Experiments}

In this section, we conduct extensive experiments to validate the effectiveness of our proposed method. We first introduce the evaluation benchmark and dataset used in our experiments. Then, through ablation studies and qualitative visualizations, we analyze the contribution of each key component in our framework. Finally, we demonstrate the generality of our approach by integrating it with other advanced driving models, showing consistent performance improvements.

\subsection{Benchmark}

\textbf{NAVSIM v1.~\cite{navsim}} This is a data-driven non-reactive simulation benchmark that bridges the gap between open-loop and closed-loop evaluation. NAVSIM unrolls bird's-eye-view abstractions of real-world scenes over a short simulation horizon to compute simulation-based metrics such as progress and time-to-collision, offering better alignment with closed-loop performance than traditional displacement errors while maintaining computational efficiency. The benchmark is built upon the OpenScene dataset (a redistribution of nuPlan~\cite{nuplan}) and provides standardized filtered splits to focus on challenging driving scenarios. The dataset is split into two subsets: \textbf{navtrain} with approximately 85k scenes and \textbf{navtest} with around 12k scenes. The evaluation metric, PDM Score (\textbf{PMDS}),
 is a weighted combination of five sub-metrics:  no collisions (NC), drivable area compliance (DAC), time-to-collision (TTC), comfort (Comf.), and Ego Progress (EP). The formal computation is defined as:
\begin{equation}
\mathrm{PDMS} = \mathrm{NC}\times \mathrm{DAC}\times \frac{5\cdot \mathrm{EP}+5\cdot \mathrm{TTC}+2\cdot \mathrm{Comf.}}{12},
\end{equation} 

\noindent \textbf{NAVSIM v2.~\cite{navsimv2}} Building upon v1, this version introduces a reactive simulation environment where background traffic agents dynamically respond to the ego vehicle's behavior, enabling more realistic evaluation of interactive driving scenarios . To comprehensively assess performance in this reactive setting, NAVSIM v2 employs the Extended PDM Score (\textbf{EPDMS}), which expands the metric set from five to nine sub-metrics . The multiplier metrics are extended with Driving Direction Compliance (DDC) and Traffic Light Compliance (TLC) , while the weighted metrics are augmented with Lane Keeping (LK) , History Comfort (HC) (an improved version of the basic Comfort metric), and Extended Comfort (EC).

\noindent \textbf{HUGSIM.~\cite{hugsim}} HUGSIM is a real-time, photo-realistic closed-loop simulator that lifts captured 2D RGB images into 3D space via 3D Gaussian Splatting, enabling dynamic updates of ego and actor states based on control commands. Unlike log-replay or reactive-agent benchmarks, HUGSIM supports fully interactive closed-loop evaluation: the ego agent's decisions influence the environment in real time, and surrounding vehicles are simulated reactively. \textbf{HD-Score} is used for closed-loop evaluation, It aggregates Route Completion (RC) with NC, DAC,
 TTC, Comf. across an episode of length T:
\begin{equation}
\mathrm{HD-Score} = \mathrm{RC}\times \frac{1}{T}\sum_{t=1}^{T} \left( \mathrm{NC}_t \times \mathrm{DAC}_t \times \frac{5\cdot \mathrm{TTC}_t + 2\cdot  \mathrm{Comf.}_t}{7} \right).
\end{equation}

\begin{table}[tb]
\caption{Performance on the NAVSIMv1 navtest Benchmark. "C" dontes Camera, and "L" denotes Lidar. * Indicates the optimized version.
  }
  \label{tab:navsimv1}
  \centering
  \begin{tabular}{l|cc|ccccc|c}
    \toprule
    
    {Method} & 
    Venue & 
    Input & 
    {NC$\uparrow$} & 
    {DAC$\uparrow$} & 
    {EP$\uparrow$} & 
    {TTC$\uparrow$} & 
    {Comf.$\uparrow$} &  
    {PDMS$\uparrow$} \\
    \midrule
    
    Hydra-MDP~\cite{hydra-mdp} & 
    arXiv'24 & C\&L & 98.3 & 96.0 & 78.7 & 94.6 & \textbf{100.0} & 86.5\\
    
    DiffusionDrive~\cite{diffusiondrive} & 
    CVPR'25 & C\&L & 98.2 & 96.2 & 82.2 & 94.7& \textbf{100.0} & 88.1\\
    
    WoTE~\cite{wote} & 
    ICCV'25 & C\&L & 98.5 & 96.8 & 81.9 & 94.4 & 99.9 & 88.3\\

    Hydra-NeXt~\cite{hydranext} & 
    ICCV'25 & C\&L & 98.1 & 97.7 & 81.8 & 94.6 & \textbf{100.0} & 88.6\\

    \cellcolor{lightgray}TransFuser~\cite{transfuser} & 
    \cellcolor{lightgray}CVPR'21 & 
    \cellcolor{lightgray}C\&L & 
    \cellcolor{lightgray}97.7 & 
    \cellcolor{lightgray}92.8 & 
    \cellcolor{lightgray}79.2 & 
    \cellcolor{lightgray}92.8 & 
    \cellcolor{lightgray}\textbf{100.0} & 
    \cellcolor{lightgray}84.0 \\
    
    \cellcolor{lightblue}TransFuser v7 & 
    \cellcolor{lightblue}- & 
    \cellcolor{lightblue}C\&L & 
    \cellcolor{lightblue}98.7  & 
    \cellcolor{lightblue}97.5  & 
    \cellcolor{lightblue}83.2  & 
    \cellcolor{lightblue}95.5  & 
    \cellcolor{lightblue}\textbf{100.0}  & 
    \cellcolor{lightblue}{89.6}\\  
    \midrule
    
    UniAD~\cite{uniad} & 
    CVPR'23 & C & 97.8 & 91.9 & 78.8 & 92.9 &\textbf{100.0} & 83.4\\
    
    PARA-Drive~\cite{paradrive} & 
    CVPR'24 & C & 97.9 & 92.4 & 79.3 & 93.0 & 99.8 & 84.0 \\
    
    LAW~\cite{law} & 
    ICLR'25 & C & 96.4 & 95.4 & 81.7 & 88.7 & 99.9 & 84.6\\
    
    Epona~\cite{epona} & 
    ICCV'25 & C & 97.9 & 95.1 & 80.4 & 93.8 & 99.9 & 86.2\\
    
    DrivingGPT~\cite{drivinggpt} & 
    ICCV'25 & C &{98.9} & 90.7 & 79.7& 94.9&95.6 & 82.4\\ 
    
    PWM~\cite{pwm} & 
    NeurIPS'25 & C & 98.6 & 95.9 & 81.8 & 95.4 & \textbf{100.0} & 88.1\\
    
    AutoVLA~\cite{autovla} & 
    NeurIPS'25 & C & 98.4 & 95.6 & 81.9 & 98.0 & 99.9 & 89.1\\

    WorldRFT~\cite{worldrft} & 
    AAAI'26 & C & 97.8 & 96.8 & 81.7 & 94.0 & \textbf{100.0} & 87.8\\

    VADv2~\cite{vadv2} & ICLR'26 & 
    C & 98.3 & 97.4 & 82.3 & 95.7 & \textbf{100.0} & 89.3\\

    BridgeDrive~\cite{bridgedrive} & 
    ICLR'26 & C & 98.2 & 96.1 & 82.3 & 94.5 & \textbf{100.0} & 88.0\\

    SNG-VLA~\cite{hua2026unveiling} & 
    ICRA'26 & C & 98.9 & 96.5 & \textbf{83.8} & 92.9 & \textbf{100.0} & 88.2\\
    
    VGGDrive\cite{vggdrive} & 
    CVPR'26 & C & 98.6 & 96.3 & 82.9 & {95.6} & 99.9 & 88.8\\

    DriveLaW\cite{drivelaw} & 
    CVPR'26 & C & \textbf{99.0} & 97.1 & 81.3 & \textbf{96.7} & \textbf{100.0} & 89.1\\
    MeanFuser\cite{meanfuser} & 
    CVPR'26 & C & 98.6 & 97.0 & 82.8 & 95.0 & \textbf{100.0} & 89.0\\    
    \cellcolor{lightgray}LTF~\cite{transfuser} & 
    \cellcolor{lightgray}NeurIPS'24 & 
    \cellcolor{lightgray}C & 
    \cellcolor{lightgray}97.4 & 
    \cellcolor{lightgray}92.8 & 
    \cellcolor{lightgray}79.0 & 
    \cellcolor{lightgray}92.4 & 
    \cellcolor{lightgray}\textbf{100.0} & 
    \cellcolor{lightgray}83.8\\
    
    \cellcolor{lightgray}LTFv6~\cite{lead} & 
    \cellcolor{lightgray}CVPR'26 & 
    \cellcolor{lightgray}C & 
    \cellcolor{lightgray}97.5 & 
    \cellcolor{lightgray}95.4 & 
    \cellcolor{lightgray}80.9 & 
    \cellcolor{lightgray}93.8 & 
    \cellcolor{lightgray}\textbf{100.0} & 
    \cellcolor{lightgray}86.4\\

    \cellcolor{lightgray}LTF* & 
    \cellcolor{lightgray}- & 
    \cellcolor{lightgray}C & 
    \cellcolor{lightgray}98.2 & 
    \cellcolor{lightgray}97.0 & 
    \cellcolor{lightgray}83.0 & 
    \cellcolor{lightgray}94.6 & 
    \cellcolor{lightgray}\textbf{100.0} & 
    \cellcolor{lightgray}88.7\\
    
    \cellcolor{lightblue}LTFv7 & 
    \cellcolor{lightblue}- & 
    \cellcolor{lightblue}C & 
    \cellcolor{lightblue}98.4  & 
    \cellcolor{lightblue}\textbf{97.9}  & 
    \cellcolor{lightblue}{83.7}  & 
    \cellcolor{lightblue}95.0 & 
    \cellcolor{lightblue}\textbf{100.0} & 
    \cellcolor{lightblue}\textbf{89.7} \\  
    \bottomrule
  \end{tabular}
\end{table}

\begin{table}[tb]
\caption{Performence on the NAVSIMv2 navtest Benchmark.
  }
  \label{tab:navsimv2}
  \centering
  \begin{tabular}{l|cccc|ccccc|c}
    \toprule
    
    {Method} & 
    {NC$\uparrow$} & 
    {DAC$\uparrow$} & 
    {DDC$\uparrow$} & 
    {TLC$\uparrow$} & 
    {EP$\uparrow$} & 
    {TTC$\uparrow$} & 
    {LK$\uparrow$} & 
    {HC$\uparrow$} & 
    {EC$\uparrow$} &  
    {EPDMS$\uparrow$}  \\
    \midrule
    
    DriveSuprim~\cite{drivesuprim} & 
    97.5 & 96.5 & 99.4 & 99.6 & \textbf{88.4} & 96.6 & 95.5 & \textbf{98.3} & 77.0 & 83.1\\
    
    ARTEMIS~\cite{artemis} & 
    98.3 & 95.1 & 98.6 & \textbf{99.8} & 81.5 & 97.4 & 96.5 & - & \textbf{98.3} & 83.1\\
    
    DiffusionDrive~\cite{diffusiondrive} & 
    98.2 & 96.3 & 99.4 & \textbf{99.8} & 87.4 & 97.4 & 97.0 & \textbf{98.3} & 87.7 & 88.3\\
    
    MeanFuser~\cite{meanfuser} & 
    98.3 & 97.4 & \textbf{99.6} & \textbf{99.8} & {87.5} & 97.6 & \textbf{97.5} & \textbf{98.3} & 88.0 & 89.6\\
    \midrule 
    \cellcolor{lightgray}TransFuser~\cite{transfuser} & 
    \cellcolor{lightgray}97.7 & \cellcolor{lightgray}92.8 & 
    \cellcolor{lightgray}99.2 & 
    \cellcolor{lightgray}99.7 & 
    \cellcolor{lightgray}87.5 & 
    \cellcolor{lightgray}96.5 & 
    \cellcolor{lightgray}95.9 & 
    \cellcolor{lightgray}\textbf{98.3} & 
    \cellcolor{lightgray}88.3 & 
    \cellcolor{lightgray}84.4 \\

    \cellcolor{lightblue}TransFuser v7 & 
    \cellcolor{lightblue}{98.3} & 
    \cellcolor{lightblue}\textbf{98.0} & 
    \cellcolor{lightblue}{99.5} & 
    \cellcolor{lightblue}\textbf{99.8} & 
    \cellcolor{lightblue}87.3 & 
    \cellcolor{lightblue}97.5 & 
    \cellcolor{lightblue}{97.3} & 
    \cellcolor{lightblue}\textbf{98.3} & 
    \cellcolor{lightblue}88.3 & 
    \cellcolor{lightblue}90.0 \\ 

    \cellcolor{lightgray}LTF~\cite{navsim} & 
    \cellcolor{lightgray}97.6 & \cellcolor{lightgray}91.9 & 
    \cellcolor{lightgray}99.2 & 
    \cellcolor{lightgray}\textbf{99.8} & 
    \cellcolor{lightgray}87.6 & 
    \cellcolor{lightgray}97.2 & 
    \cellcolor{lightgray}96.6 & 
    \cellcolor{lightgray}\textbf{98.3} & 
    \cellcolor{lightgray}86.3 & 
    \cellcolor{lightgray}83.6 \\

    \cellcolor{lightgray}LTF* & 
    \cellcolor{lightgray}98.3 & \cellcolor{lightgray}97.0 & 
    \cellcolor{lightgray}99.5 & 
    \cellcolor{lightgray}\textbf{99.8} & 
    \cellcolor{lightgray}87.5 & 
    \cellcolor{lightgray}97.3 & 
    \cellcolor{lightgray}97.3 & 
    \cellcolor{lightgray}\textbf{98.3} & 
    \cellcolor{lightgray}88.2 & 
    \cellcolor{lightgray}89.1 \\
    
    \cellcolor{lightblue}LTFv7 & 
    \cellcolor{lightblue}\textbf{98.4} & 
    \cellcolor{lightblue}{97.9} & 
    \cellcolor{lightblue}{99.5}  & 
    \cellcolor{lightblue}\textbf{99.8}  & 
    \cellcolor{lightblue}{87.8}  & 
    \cellcolor{lightblue}\textbf{98.0}  & 
    \cellcolor{lightblue}{97.3} & 
    \cellcolor{lightblue}\textbf{98.3}  & 
    \cellcolor{lightblue}\textbf{88.5} & 
    \cellcolor{lightblue}\textbf{90.1}\\
    
    \bottomrule
  \end{tabular}
\end{table}

\begin{table}[!t]
\centering

\caption{Zero-shot Performance on the HUGSIM Benchmark.}
\resizebox{\textwidth}{!}{
\begin{tabular}{l|cccccccc|cc}
\toprule
\multirow{2}{*}{Method} 
& \multicolumn{2}{c}{Easy} 
& \multicolumn{2}{c}{Medium} 
& \multicolumn{2}{c}{Hard} 
& \multicolumn{2}{c|}{Extreme} 
& \multicolumn{2}{c}{Overall} \\
\cmidrule{2-3} \cmidrule{4-5} \cmidrule{6-7} \cmidrule{8-9} \cmidrule{10-11}
 & RC & HD-Score & RC & HD-Score & RC & HD-Score & RC & HD-Score & RC & HD-Score \\
\midrule

UniAD~\cite{uniad}
& 58.6 & 48.7 &  41.2 &  29.5 & \textbf{40.4}  & \textbf{27.3}  & 26.0  &  14.3 & 40.6  &  28.9 \\

VAD~\cite{vad}
& 38.7 & 24.3 & 27.0  & 9.9  & 25.5  & 10.4  & 23.0  & 8.2  &  27.9 & 12.3  \\

MeanFuser~\cite{meanfuser}
& 76.7 & \textbf{67.1} & 41.1  & 24.3  & 37.9  & 21.5  & 28.4  & 11.0  &  27.9 & 12.3  \\
\midrule

\cellcolor{lightgray}LTF~\cite{navsim} & 
\cellcolor{lightgray}68.4 & 
\cellcolor{lightgray}52.8 & 
\cellcolor{lightgray}40.7 & 
\cellcolor{lightgray}24.6 & 
\cellcolor{lightgray}36.9 &  
\cellcolor{lightgray}19.8 & 
\cellcolor{lightgray}25.5 &  
\cellcolor{lightgray}8.1 &  
\cellcolor{lightgray}41.4 &  
\cellcolor{lightgray}24.8 \\

\cellcolor{lightblue}LTFv7 & 
\cellcolor{lightblue}\textbf{76.8}  & 
\cellcolor{lightblue}{65.6} &  
\cellcolor{lightblue}\textbf{43.0} &  
\cellcolor{lightblue}\textbf{31.4} & 
\cellcolor{lightblue}35.5 & 
\cellcolor{lightblue}26.3  &  
\cellcolor{lightblue}\textbf{29.6} & 
\cellcolor{lightblue}\textbf{16.2}  &  
\cellcolor{lightblue}\textbf{46.2} &  
\cellcolor{lightblue}\textbf{34.8} \\
\bottomrule
\end{tabular}
}
\label{tab:hugsim}
\vspace{-0.4cm}
\end{table}

\begin{table}[tb]
\caption{Ablation studies on core components for both uni-modal and multi-modal methods. Hyperparameter Tuning includes replacing the constant scheduler with a cosine annealing scheduler with warmup, as well as adjusting the loss weights. Normalization is performed by using first-order differences of coordinates instead of raw coordinates.
  }
  
  \label{tab:ablation_module}
  \centering
  \resizebox{\textwidth}{!}{
  \begin{tabular}{l|l|ccccc|l|l}
    \toprule
    
    {Model} & 
    Components & 
    {NC$\uparrow$} & 
    {DAC$\uparrow$} & 
    {EP$\uparrow$} & 
    {TTC$\uparrow$} & 
    {Comf.$\uparrow$} & 
    {PDMS$\uparrow$} & 
    {EPDMS$\uparrow$} \\
    \midrule

    \rowcolor[HTML]{EEE5F4}
    \multicolumn{9}{c}{\textit{{Uni-Modal Planner}}} \\
    
    Tranfsuer\cite{transfuser} & - & 97.7 & 92.8 & 79.2 & 92.8 & 100 & 84.0 & 84.4 \\
    
    LTF\cite{navsim} & Baseline & 97.4 & 92.8 & 79.0 & 92.4 & 100 & 83.8 & 83.6 \\
    
    \(\mathcal{M}_{s1}\) & LTF + Hyperparameter Tuning & 98.1 & 95.7 & 81.7 & 94.3 & 100 & 87.4 & 87.8 \\
    
    LTF* & 
    \(\mathcal{M}_{s1}\) + Trajectory Normalization & 98.2 & 97.0 & 83.0 & 94.6 & 100 & 88.7 & 89.1 \\
    
    \cellcolor{lightgray}LTFv7 & 
    \cellcolor{lightgray}LTF* + BeyondDrive & \cellcolor{lightgray}98.4 & 
    \cellcolor{lightgray}97.9 & 
    \cellcolor{lightgray}83.7 & 
    \cellcolor{lightgray}95.0 & 
    \cellcolor{lightgray}100 & \cellcolor{lightgray}89.7$_{\textcolor{green!50!black}{+1.0}}$ & \cellcolor{lightgray}90.1$_{\textcolor{green!50!black}{+1.0}}$\\
    \midrule 

    \rowcolor[HTML]{EEE5F4}
    \multicolumn{9}{c}{\textit{Diffusion-based Planner}} \\

    DiffusionDrive\cite{diffusiondrive} & 
    Baseline & 
    98.2 & 
    96.2 & 
    82.2 & 
    94.7 & 
    100 & 
    88.1 & 
    88.3 \\

    \cellcolor{lightgray}\(\mathcal{M}_{m1}\) & 
    \cellcolor{lightgray}DiffusionDrive + BeyondDrive & 
    \cellcolor{lightgray}98.2 & 
    \cellcolor{lightgray}97.4 & 
    \cellcolor{lightgray}83.4 & 
    \cellcolor{lightgray}94.8 & 
    \cellcolor{lightgray}100 & 
    \cellcolor{lightgray}89.2$_{\textcolor{green!50!black}{+1.1}}$ & 
    \cellcolor{lightgray}88.9$_{\textcolor{green!50!black}{+0.6}}$ \\ 
    \midrule 

    \rowcolor[HTML]{EEE5F4}
    \multicolumn{9}{c}{\textit{{Flow-based Planner}}} \\

    MeanFuser\cite{meanfuser} & 
    Baseline & 
    98.6 & 
    97.0 & 
    82.8 & 
    95.0 & 
    100 & 
    89.0 & 
    89.6 \\

    \cellcolor{lightgray}\(\mathcal{M}_{m2}\) & 
    \cellcolor{lightgray}MeanFuser + BeyondDrive & 
    \cellcolor{lightgray}98.5 & 
    \cellcolor{lightgray}98.4 & 
    \cellcolor{lightgray}84.2 & 
    \cellcolor{lightgray}95.1 & 
    \cellcolor{lightgray}100 & 
    \cellcolor{lightgray}90.3$_{\textcolor{green!50!black}{+1.3}}$ & 
    \cellcolor{lightgray}90.5$_{\textcolor{green!50!black}{+0.9}}$ \\ 
    \midrule 

    \rowcolor[HTML]{EEE5F4}
    \multicolumn{9}{c}{\textit{{World Model Planner}}} \\

    WoTE\cite{wote} & 
    Baseline & 
    98.5 & 
    96.8 & 
    81.9 & 
    94.4 & 
    99.9 & 
    88.3 & 
    87.6 \\

    \cellcolor{lightgray}\(\mathcal{M}_{w1}\) & 
    \cellcolor{lightgray}WoTE + BeyondDrive & 
    \cellcolor{lightgray}98.4 & 
    \cellcolor{lightgray}97.3 & 
    \cellcolor{lightgray}83.1 & 
    \cellcolor{lightgray}95.1 & 
    \cellcolor{lightgray}100 & 
    \cellcolor{lightgray}89.2$_{\textcolor{green!50!black}{+0.9}}$ & 
    \cellcolor{lightgray}88.7$_{\textcolor{green!50!black}{+1.1}}$ \\ 
    
    \bottomrule
  \end{tabular}}
\end{table}

\subsection{Implementation Details}

We trained all models with a batch size of 32 for a total of 100 epochs. We employed a cosine annealing learning rate scheduler with three warmup epochs and a peak learning rate of 2e-4. The Flow Matching model uses the exact same perception module as LTF~\cite{navsim}. For negative sample generation, we used a classifier-free guidance (CFG) scale of 0.5 and 5 sampling steps in the flow matching process. 

\subsection{Main Results}
\textbf{Results on NAVSIM.} We train two variants of our model under the same setting: \textbf{TransFuser v7}, which takes both camera and point cloud inputs, and \textbf{LTFv7}, which uses camera inputs only. As shown in ~\cref{tab:navsimv1}, our method achieves promising performance in both settings, with PDMS scores of 89.6 and 89.7, respectively, surpassing many recent high-performing multimodal evaluators\cite{vadv2,hydra-mdp,diffusiondrive,hydranext,meanfuser}, world models\cite{law,wote,epona,pwm,drivelaw}, and Vision-Language-Action (VLA) based\cite{drivinggpt,autovla,vggdrive} approaches. The comparable performance between TransFuser v7 and LTFv7 is consistent with the official NAVSIM benchmark results. Moreover, compared to LTF, LTFv7 improves PDMS by \textbf{5.9}, with NC and DAC increasing by \textbf{1.0} and \textbf{5.1}, respectively. These gains indicate a significant reduction in collisions and drivable area departures, while also enhancing traffic efficiency (EP \textbf{+4.7}) and maintaining a safer distance (TTC \textbf{+2.6}). \Cref{tab:navsimv2} reports results on the reactive closed-loop benchmark NAVSIM v2, where both TransFuser v7 and LTFv7 exceed 90 EPDMS. Similarly, substantial gains are observed in the DAC metric, and the improved LK (\textbf{+0.7}) metric demonstrates that our approach enhances long-range planning capability.

\textbf{Results on HUGSIM.} As shown in \cref{tab:hugsim}, we evaluate the zero-shot closed-loop performance of our method on four official datasets (KITTI-360\cite{kitti-360}, Waymo\cite{waymo}, nuScenes\cite{nuscenes}, and Pandaset\cite{pandaset}). Our approach achieves state-of-the-art results in the levels of difficulty of easy, Medium, and extreme. Compared to LTF, it improves the overall Rc and HD-Score by \textbf{+4.8} and \textbf{+10}, respectively, demonstrating improved driving performance and safety while exhibiting a strong generalizability. Notably, consistent improvements are also observed on metrics such as DDC and LK within EPDMS, which are not present in the PDMS metric used for negative sample selection. This further confirms that our method learns a transferable notion of safety beyond mere optimization of a single simulator metric, and that the performance gains are not contingent on the specific scoring function used during training. 

\begin{table}[!h]
\caption{Ablation on key hyperparameters. \(N\): number of flow matching samples per scene. \(w\): CFG scale. \(\sigma_{\text{sample}}\): noise scale at sampling. \(\mathrm{Prop}_{\text{sample}}\): proportion of scenes with valid negative samples. \(\mathrm{PDMS}_{\text{max}}\) and \(\mathrm{PDMS}_{\text{std}}\): average (over scenes) of the maximum and standard deviation of PDMS across \(N\) trajectories. \(\mathrm{Dist}_{\text{nega}}\) and \(\mathrm{PDMS}_{\text{nega}}\): average distance with expert demonstration and average PDMS of the selected negatives. \(\mathrm{PDMS}_{\text{LTF}}\): PDMS of the LTF baseline.}
  
  \label{tab:ablation_flow}
  \centering
  \resizebox{\textwidth}{!}
  {
  \begin{tabular}{ccc|ccc|cc|c}
    \toprule
    
    \(N\) & 
    \(w\) & 
    \(\sigma_{sample}\) & 
    \(\mathrm{Prop}_{\mathrm{sample}}\) & 
    \(\mathrm{PDMS}_{\mathrm{max}}\) & 
    \(\mathrm{PDMS}_{\mathrm{std}}\) & 
    \(\mathrm{Dist}_{\mathrm{nega}}\) & 
    \(\mathrm{PDMS}_{\mathrm{nega}}\) &
    \(\mathrm{PDMS}_{\mathrm{LTF}}\)
    \\
    \midrule
    
    1 & 1.0 & \(1.0\sigma\) & - & 87.1 & 0 & - & - & - \\
    
    64 & 1.0 & \(1.0\sigma\) & 8.0\% & 94.8 & 0.0008 & 0.40 & 23.9 & 88.8 \\
    
    64 & 0.5 & \(1.0\sigma\) & 78.8\% & 98.2 & 0.1900 & 1.83 & 22.8 & 89.1 \\
    
    64 & 0.5 & \(2.0\sigma\) & 96.4\% & 98.9 & 0.3392 & 1.68 & 19.8 & 89.7 \\
    
    \bottomrule
  \end{tabular}}
\end{table}

\subsection{Ablation Studies and Case Studies}

We ablate the key components and hyperparameters to investigate their impact on model performance.

\textbf{ The ablation of each core module.}
As shown in ~\cref{tab:ablation_module}, we ablate the key components that contribute to the performance improvement of the uni-modal method LTFv7. With hyperparameter tuning alone, \ie, replacing the official constant learning rate scheduler with a cosine annealing scheduler with warmup and reducing the loss weights of auxiliary tasks, the model achieves a performance gain of +3.6 PDMS. After applying normalization, the model further improves by +1.3 PDMS, as the smoothed data distribution eases network learning. Finally, incorporating BeyondDrive boosts the model to 89.7 PDMS, with EPDMS exceeding 90 and clear improvements in safety metrics such as NC, DAC, and TTC, validating the effectiveness of our method. We also applied BeyondDrive to the multimodal methods DiffusionDrive and MeanFuser, as well as the world model method WoTE. In all three methods, simply by adding one line of training loss code, we achieved PDMS gains of +1.1, +1.3, and +0.9 respectively, demonstrating the generalizability of our approach.

\textbf{The ablation of key hyperparameters.} As shown in ~\cref{tab:ablation_flow}, when \(N=1\), the flow matching model achieves a \(\mathrm{PDMS}_{\text{max}}\) of 87.1, reflecting its baseline capability. Without CFG and noise scaling (\(w=1.0\), \(\sigma_{\text{sample}}=1.0\sigma\)), only 8.0\% of scenes yield valid negative samples. Reducing the CFG scale to \(w=0.5\) substantially increases this proportion to 78.8\%. Further enlarging the sampling noise to \(\sigma_{\text{sample}}=2.0\sigma\) pushes the proportion to 96.4\%, while the average distance to the expert demonstration (\(\mathrm{Dist}_{\text{nega}}\)) drops to 1.68, and the LTF performance improves to 89.7 PDMS. These results highlight the importance of both CFG and noise scaling in generating diverse and proximal hard negatives for safety-aware learning.

\begin{figure}[tb]
  \centering
  \includegraphics[height=9.0cm]{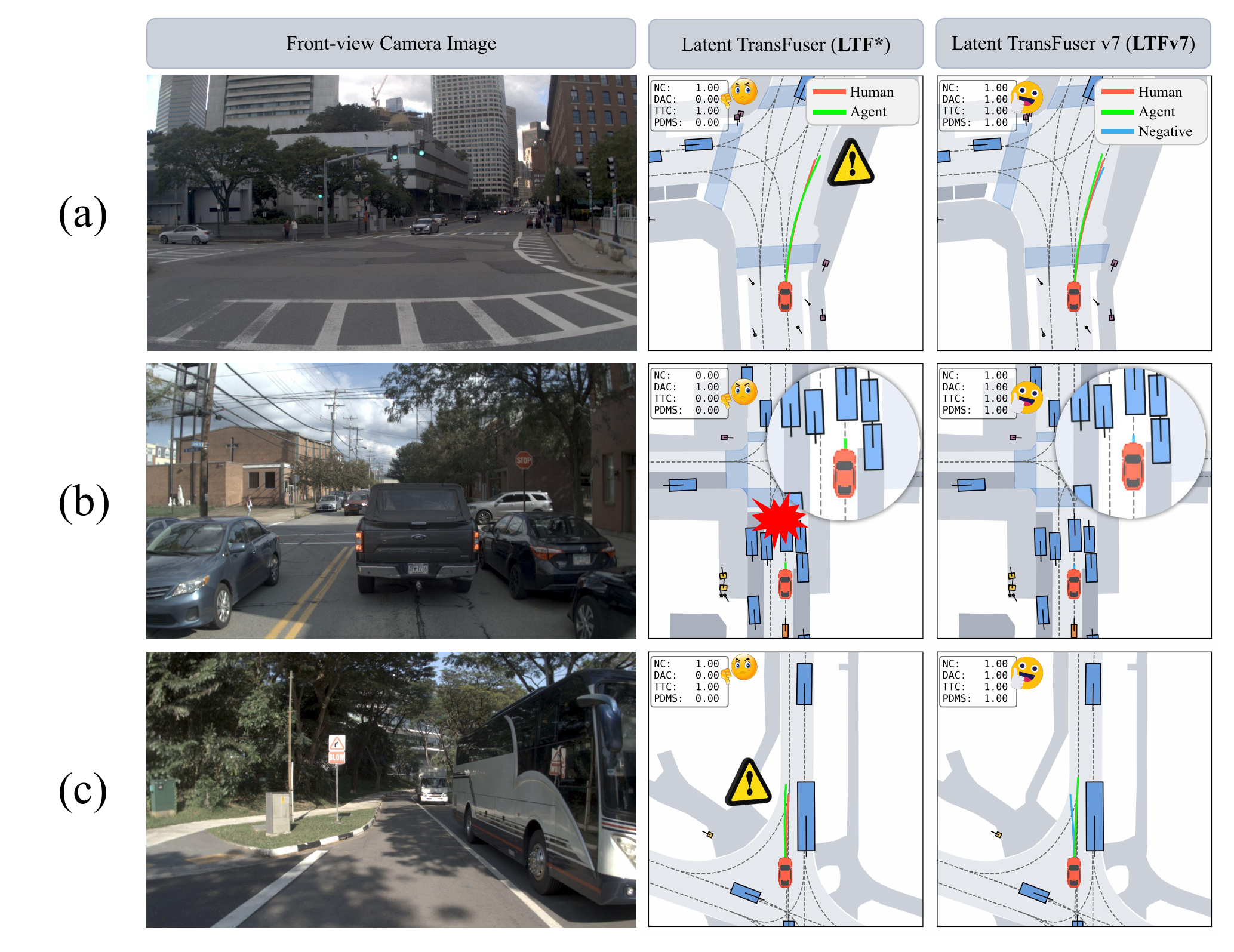}
  \caption{Case studies on the navtest benchmark. Left: front-view camera images. Middle: BEV scenes with expert demonstrations and LTF* planned trajectories (with scores). Right: expert demonstrations, LTFv7 planned trajectories and negative samples.
  }
  \label{fig:scene_nage_paper}
  \vspace{-0.4cm}
\end{figure}

\textbf{The ablation of loss weights.} We ablate the influence of the RD loss weight \(\lambda_{\text{rd}}\) on model performance. As shown in \cref{fig:loss_weights}, gradually increasing \(\lambda_{\text{rd}}\) yields significant performance improvements, provided that its value remains lower than \(\lambda_{\text{imi}}\); otherwise, the model fails to converge. More details are presented in \cref{tab:ablation_loss_weight}.

\begin{wrapfigure}{l}{0.5\textwidth}  
  \centering
  \includegraphics[width=\linewidth]{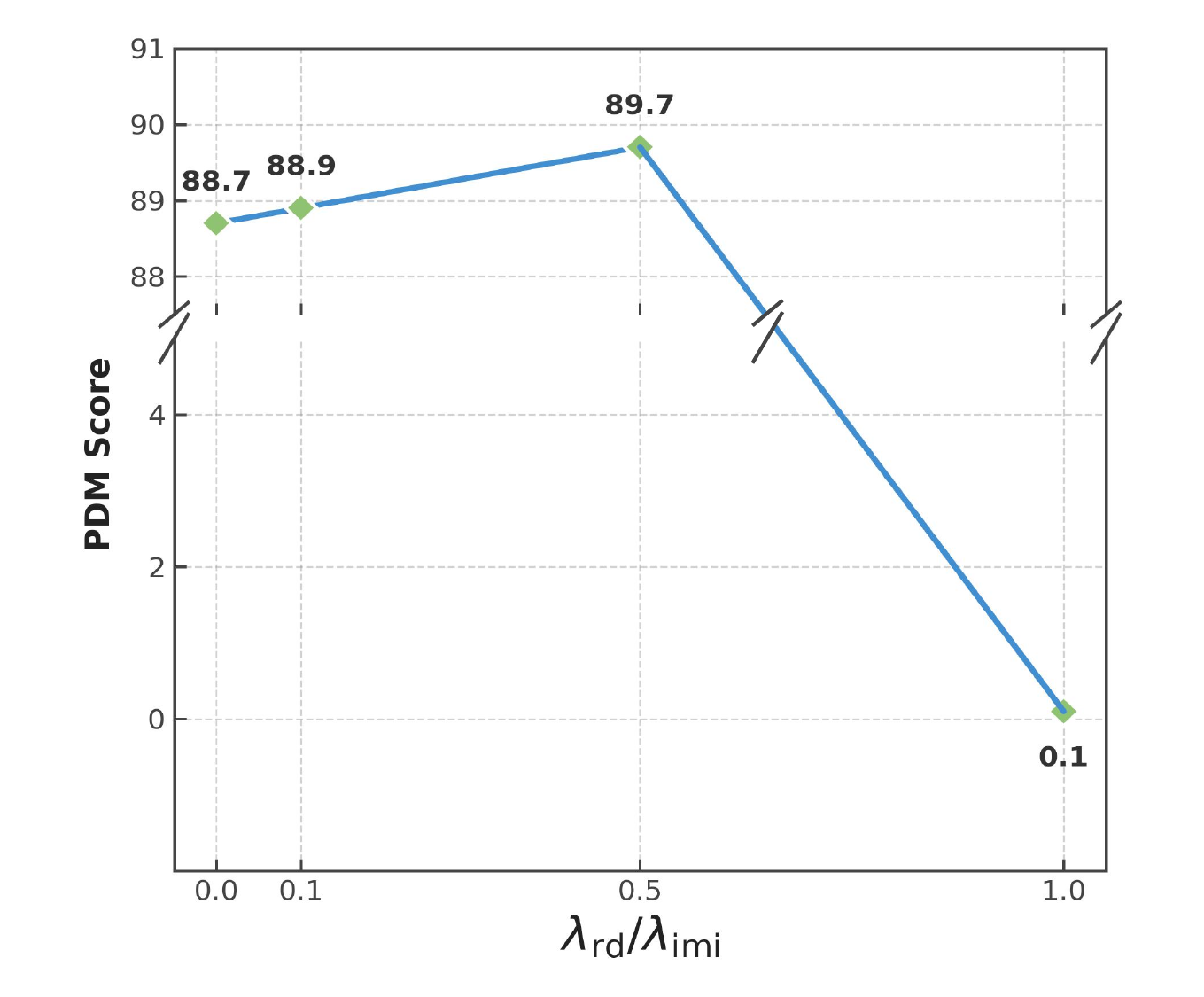}
  \caption{Impact of $\lambda_{\text{rd}}/\lambda_{\text{imi}}$ on model performance.}
  \label{fig:loss_weights}
  \vspace{-0.4cm}
\end{wrapfigure}

\textbf{Case Studies.} Specifically, we present three representative scenarios in \cref{fig:scene_nage_paper} to demonstrate the improvements of LTFv7 over LTF* in terms of planning quality and safety. (a) At a T-shaped intersection, the trajectory planned by LTF* deviates from the drivable area at the far end, resulting in a DAC score of 0. Our generated negative samples indicate that such maneuvers are unsafe, which guides the optimized LTFv7 to learn a safer driving strategy that steers clear of the road boundary. (b) At a crossroad requiring a stop, LTF* plans a forward-moving trajectory, leading to NC and TTC scores of 0, indicating an imminent collision with the front vehicle. Our negative samples demonstrate the danger of proceeding, and thus LTFv7 maintains a stationary policy. (c) In a bus-merging scenario, LTF* adopts an overly cautious yielding strategy that steers off the road edge. Corrected by the negative samples, the model learns the proper planning policy of staying near the lane center.

\section{Conclusion}
We presented BeyondDrive, a failure-aware contrastive learning framework for end-to-end autonomous driving that enforces safety boundaries via expert-proximate negative trajectories. By synthesizing unsafe yet expert-proximate trajectories with flow matching and optimizing a contrastive distance-based objective, BeyondDrive enables policies to preserve imitation fidelity while explicitly avoiding safety-critical behaviors. Experiments on NAVSIM and HUGSIM show consistent improvements across uni-modal and multi-modal planners, demonstrating strong generalization and improved closed-loop driving performance.

\section{Limitations}

BeyondDrive focuses on trajectory-level safety learning and does not explicitly model long-horizon interactions or high-level semantic constraints, both of which are deferred to future work. Additionally, its integration with vocabulary-based approaches such as VADv2 is limited by the need for an additional refinement module following the scoring phase to facilitate safety-oriented training.

\newpage

\bibliographystyle{splncs04}
\bibliography{main}

\newpage
 
\section*{Supplementary Materials}
\section{Model Training and Sampling}

\begin{algorithm}[!h]
\caption{Flow Matching Generator Training with CFG}
\label{alg:training}
\KwIn{Training dataset $\mathcal{D}$, expert trajectory $\boldsymbol{\tau}^{\text{exp}}$, noise scale $\sigma$, condition dropout probability $p$, map loss weight $\lambda_{\text{map}}$, networks $\mathcal{E}_{\theta}$ and $v_{\theta}$ with initial parameters $\theta$, number of trajectory points $K$ and dimension $D$.}
\BlankLine
\For{each iteration}{
$(\boldsymbol{\tau}_{\text{exp}}, \mathbf{I}_t, s_t, c_t) \sim \mathcal{D}$; $\mathbf{x}_1 \in \mathbb{R}^{K \times D} \leftarrow \mathrm{Norm}(\boldsymbol{\tau}^{\text{exp}})$\;
$\mathcal{C} \leftarrow \mathcal{E}_{\theta}(\mathbf{I}_t, s_t, c_t)$ \tcc*{encode scene data} 
$\mathbf{x}0 \sim \mathcal{N}(\mathbf{0}, \sigma^2\mathbf{I})$, $t \sim \mathcal{U}(0,1)$, $\mathcal{C}_{\text{in}} \leftarrow \mathcal{C}$ with probability $1-p$ else $\emptyset$ \;
$\mathbf{x}_t \leftarrow (1-t)\mathbf{x}_0 + t\mathbf{x}_1$ \;
$\mathbf{v}_{\text{pred}} \leftarrow v_\theta(\mathbf{x}_t, t \mid \mathcal{C}_{\text{in}})$ \;
$\mathcal{L}_{\text{FM}} \leftarrow \| \mathbf{v}_{\text{pred}} - (\mathbf{x}_1 - \mathbf{x}_0) \|_1$ \;
$\mathcal{L} \leftarrow \mathcal{L}_{\text{FM}} + \lambda_{\text{map}} \mathcal{L}_{\text{map}}$ \tcc*{add map reconstruction loss}
Update $\mathcal{E}_{\theta}$, $v_\theta$ to minimize $\mathcal{L}$;
}
\end{algorithm}

\begin{algorithm}[!h]
\caption{Hard Negative Sampling with Flow Matching}
\label{alg:sample}
\KwIn{Trained encoder $\mathcal{E}_{\theta}$ and velocity network $v_\theta$, conditioning input $(\mathbf{I}_t, s_t, c_t)$, guidance scale $w$, sampling noise scale $\sigma_{\text{sample}}$ (e.g., $2.0\sigma$), number of candidates $N$, ODE steps $T_{step}$, score threshold $\xi$.}
\BlankLine
$\mathcal{C} \leftarrow \mathcal{E}_{\theta}(\mathbf{I}_t, s_t, c_t)$ \;
$\Delta t \leftarrow 1/T_{step}$\;
Initialize $\mathbf{X}_0 \in \mathbb{R}^{N \times K \times D}$ with $\mathbf{X}_0^{(i)} \sim \mathcal{N}(\mathbf{0}, \sigma_{\text{sample}}^2 \mathbf{I})$ for $i=1,\dots,N$\;
$\mathbf{X} \leftarrow \mathbf{X}_0$\;
\For{$n = 1$ \KwTo $T_{step}$}{
    $t \leftarrow (n - 1)\Delta t$ \tcc*{current time (start of step)}
    $\tilde{\mathbf{V}} \leftarrow v_\theta(\mathbf{X}, t|\emptyset) + w\big(v_\theta(\mathbf{X}, t|\mathcal{C}) - v_\theta(\mathbf{X}, t|\emptyset)\big)$ \;
    $\mathbf{X} \leftarrow \mathbf{X} + \tilde{\mathbf{V}} \cdot \Delta t$ \tcc*{Euler update (parallel)}
}
$\{\boldsymbol{\tau}^{(i)}\}_{i=1}^N \leftarrow \mathrm{Norm}^{-1}(\mathbf{X})$ \tcc*{recover trajectories}
$\mathcal{U}_{\text{low}} \leftarrow \{ \boldsymbol{\tau}^{(i)} \mid S(\boldsymbol{\tau}^{(i)}) < \xi \}$ \tcc*{safety scores ${S(\boldsymbol{\tau}^{(i)})}$.}
$\boldsymbol{\tau}^{\text{neg}} \leftarrow \arg\min_{\boldsymbol{\tau} \in \mathcal{U}_{\text{low}}} \| \boldsymbol{\tau} - \boldsymbol{\tau}^{\text{exp}} \|_2$ \tcc*{skip  when $\mathcal{U}_{\text{low}}=\emptyset$}
\end{algorithm}

\section{Details of the Experimental Results}

\begin{table}[!h]
    \caption{Ablation studies on core components on the NAVSIMv2 navtest Benchmark. \(\mathcal{M}_{s1}\) denotes the fine-tuned hyperparameters of the uni-modal method LTF; \(\mathcal{M}_{s2}\) represents \(\mathcal{M}_{s1}\) with normalization; LTFv7 is derived by further incorporating BeyondDrive into \(\mathcal{M}_{s2}\). \(\mathcal{M}_{m1}\) denotes the application of BeyondDrive to the multi-modal method DiffusionDrive, and \(\mathcal{M}_{m2}\) denotes the application of BeyondDrive to the multi-modal method MeanFuser. \(\mathcal{M}_{w1}\) denotes the application of BeyondDrive to the world model method WoTE. 
  }
  \label{tab:ablation_module_v2}
  \centering
  \resizebox{\textwidth}{!}{
  \begin{tabular}{l|ccccccccc|l}
    \toprule
    
    {Model} & 
    {NC$\uparrow$} & 
    {DAC$\uparrow$} & 
    {DDC$\uparrow$} & 
    {TLC$\uparrow$} & 
    {EP$\uparrow$} & 
    {TTC$\uparrow$} & 
    {LK$\uparrow$} & 
    {HC$\uparrow$} & 
    {EC$\uparrow$} & 
    {EPDMS$\uparrow$} \\
    \midrule

    \rowcolor[HTML]{EEE5F4}
    \multicolumn{11}{c}{\textit{Uni-Modal Planner}} \\
    
    Tranfsuer\cite{transfuser} & 97.7 & 92.8 & 99.2 & 99.7 & 87.5 & 96.5 & 95.9 & 98.3 & 88.3 & 94.4 \\
    
    LTF\cite{navsim} & 97.6 & 91.9 & 99.2 & \textbf{99.8} & 87.6 & 97.2 & 96.6 & \textbf{98.3} & 86.3 & 83.6 \\
    
    \(\mathcal{M}_{s1}\) &  98.1 & 95.7 & 99.4 & \textbf{99.8} & 87.5 & 97.2 &  97.2 & \textbf{98.3} & 88.1 & 87.8 \\
    
    LTF* & 98.3 & 97.0 & \textbf{99.5} & \textbf{99.8} & 87.5 & 97.3 & \textbf{97.3} & \textbf{98.3} & 88.2 & 89.1 \\
    
    \cellcolor{lightgray}LTFv7 & 
    \cellcolor{lightgray}\textbf{98.4} & 
    \cellcolor{lightgray}\textbf{97.9} & 
    \cellcolor{lightgray}\textbf{99.5} & 
    \cellcolor{lightgray}\textbf{99.8} & 
    \cellcolor{lightgray}\textbf{87.8} & 
    \cellcolor{lightgray}\textbf{98.0} & 
    \cellcolor{lightgray}\textbf{97.3} & 
    \cellcolor{lightgray}\textbf{98.3} & 
    \cellcolor{lightgray}\textbf{88.5} & \cellcolor{lightgray}\textbf{90.1}$_{\textcolor{green!50!black}{+1.0}}$\\
    \midrule 

    \rowcolor[HTML]{EEE5F4}
    \multicolumn{11}{c}{\textit{Diffusion-based Planner}} \\

    DiffusionDrive\cite{diffusiondrive} &  
    \textbf{98.2} & 
    96.3 & 
    \textbf{99.4} & 
    \textbf{99.8} & 
    87.4 & 
    97.4 & 
    \textbf{97.0} & 
    \textbf{98.3} & 
    \textbf{87.7} & 
    88.3 \\

    \cellcolor{lightgray}\(\mathcal{M}_{m1}\) &  
    \cellcolor{lightgray}98.1 & 
    \cellcolor{lightgray}\textbf{97.4} & 
    \cellcolor{lightgray}\textbf{99.4} & 
    \cellcolor{lightgray}\textbf{99.8} & 
    \cellcolor{lightgray}\textbf{87.7} & 
    \cellcolor{lightgray}\textbf{98.0} & 
    \cellcolor{lightgray}96.5 &
    \cellcolor{lightgray}\textbf{98.3} &
    \cellcolor{lightgray}84.5 & \cellcolor{lightgray}\textbf{88.9}$_{\textcolor{green!50!black}{+0.6}}$ \\ 
    \midrule 
    
    \rowcolor[HTML]{EEE5F4}
    \multicolumn{11}{c}{\textit{Flow-based Planner}} \\

    MeanFuser\cite{meanfuser} &  
    {98.3} & 
    97.4 & 
    \textbf{99.6} & 
    \textbf{99.8} & 
    \textbf{87.5} & 
    97.6 & 
    \textbf{97.5} & 
    \textbf{98.3} & 
    \textbf{88.0} & 
    89.6 \\

    \cellcolor{lightgray}\(\mathcal{M}_{m2}\) &  
    \cellcolor{lightgray}\textbf{98.5} & 
    \cellcolor{lightgray}\textbf{98.4} & 
    \cellcolor{lightgray}{99.5} & 
    \cellcolor{lightgray}\textbf{99.8} & 
    \cellcolor{lightgray}{87.4} & 
    \cellcolor{lightgray}\textbf{97.8} & 
    \cellcolor{lightgray}{97.3} &
    \cellcolor{lightgray}\textbf{98.3} &
    \cellcolor{lightgray}87.7 & \cellcolor{lightgray}\textbf{90.5}$_{\textcolor{green!50!black}{+0.9}}$ \\ 
    \midrule

    \rowcolor[HTML]{EEE5F4}
    \multicolumn{11}{c}{\textit{World Model Planner}} \\

    WoTE\cite{wote} &  
    \textbf{98.5} & 
    96.8 & 
    98.8 & 
    \textbf{99.8} & 
    86.1 & 
    97.8 & 
    95.4 &
    \textbf{98.3} &
    82.9 &
    87.6 \\

    \cellcolor{lightgray}\(\mathcal{M}_{w1}\) &  
    \cellcolor{lightgray}98.4 & 
    \cellcolor{lightgray}\textbf{97.4} & 
    \cellcolor{lightgray}\textbf{98.9} & 
    \cellcolor{lightgray}\textbf{99.8} & 
    \cellcolor{lightgray}\textbf{87.4} & 
    \cellcolor{lightgray}\textbf{97.8} & 
    \cellcolor{lightgray}\textbf{95.6} &
    \cellcolor{lightgray}\textbf{98.3} &
    \cellcolor{lightgray}\textbf{85.7} & \cellcolor{lightgray}\textbf{88.7}$_{\textcolor{green!50!black}{+1.1}}$ \\ 
    
    \bottomrule
  \end{tabular}}
\end{table}

\subsection{Ablation studies Details on NAVSIMv2 Benchmerk.}

\Cref{tab:ablation_module_v2} provides supplementary experimental details to \cref{tab:ablation_module} on the NAVSIMv2 benchmark. We present the scores of the individual sub-metrics on NAVSIMv2 and analyze the performance gains achieved by BeyondDrive across three distinct categories of methods: the uni-modal method TransFuser\cite{transfuser}, the multi-modal method DiffusionDrive\cite{diffusiondrive}, and the world model method WoTE\cite{wote}. BeyondDrive demonstrates consistent performance improvements across all these method types, underscoring the generality of our proposed approach. It is worth noting that LTFv7 achieves gains or maintains parity across all sub-metrics compared to \(\mathcal{M}_{s2}\), demonstrating that our method can comprehensively consider aspects such as safety, comfort, and efficiency. Meanwhile, \(\mathcal{M}_{m1}\) shows slight degradation on some metrics relative to DiffusionDrive, primarily due to the limitations of the trajectory selection module. On the world model method WoTE, BeyondDrive exhibits improvements on almost all metrics.

\begin{table}[!h]
    \caption{Ablation studies on different loss weights.
  }
  \label{tab:ablation_loss_weight}
  \centering
  \scalebox{1.1}{
  \begin{tabular}{c|ccc|ccccc|l}
    \toprule
    Setting &
    {\(\lambda_{\text{imi}}\)} & 
    {\(\lambda_{\text{sem}}\)} & 
    {\(\lambda_{\text{rd}}\)} & 
    {NC$\uparrow$} & 
    {DAC$\uparrow$} & 
    {EP$\uparrow$} & 
    {TTC$\uparrow$} & 
    {Comf.$\uparrow$} & 
    {PDMS$\uparrow$} \\
    \midrule
    
    \( \mathcal{S}_0\) & 1.0 & 1.0 & 1.0 & 91.1 & 10.4 & 0.5 & 88.6 & 4.6 & 3.7 \\
    \( \mathcal{S}_1\) & 5.0 & 1.0 & 1.0 & 98.1 & 97.0 & 82.8 & 94.3 & 100 & 88.6$_{\textcolor{green!50!black}{+84.9}}$ \\
    \( \mathcal{S}_2\) & 10.0 & 1.0 & 1.0 & 98.1 & 97.3 & 83.4 & 94.4 & 100 & 88.9$_{\textcolor{green!50!black}{+0.3}}$ \\
    \midrule
    \( \mathcal{S}_3\) & 10.0 & 0.0 & 0.0 & 97.9 & 96.0 & 81.5 & 93.8 & 100 & 87.2 \\
    \( \mathcal{S}_4\) & 10.0 & 1.0 & 0.0 & 98.2 & 97.0 & 83.0 & 94.6 & 100 & 88.7$_{\textcolor{green!50!black}{+1.5}}$ \\
    \( \mathcal{S}_5\) & 10.0 & 5.0 & 0.0 & 98.2 & 96.7 & 82.9 & 94.5 & 100 & 88.4$_{\textcolor{red}{-0.3}}$ \\
    \midrule
    \( \mathcal{S}_4\) & 10.0 & 1.0 & 0.0 & 98.2 & 97.0 & 83.0 & 94.6 & 100 & 88.7 \\
    \( \mathcal{S}_2\) & 10.0 &  1.0 & 1.0 & 98.3 & 97.0 & 83.1 & 94.6 & 100 & 88.9$_{\textcolor{green!50!black}{+0.2}}$ \\
    \( \mathcal{S}_6\) & 10.0 & 1.0 & 5.0 & 98.4 & 97.9 & 83.7 & 95.0 & 100 & 89.7$_{\textcolor{green!50!black}{+0.8}}$ \\
    \( \mathcal{S}_7\) & 10.0 & 1.0 & 10.0 & 83.6 & 0.3 & 0.0 & 78.8 & 0.0 & 0.1$_{\textcolor{red}{-89.6}}$ \\

    \bottomrule
  \end{tabular}}
\end{table}

\subsection{Ablation studies on Loss Weights.}

As shown in \cref{tab:ablation_loss_weight}, we analyze the impact of different loss weights in LTFv7 on performance, including the auxiliary supervision loss $\mathcal{L}_{\text{sem}}$, the expert demonstration imitation learning loss $\mathcal{L}_{\text{imi}}$, and the RD loss $\mathcal{L}_{\text{rd}}$. 

We observe that when ${\lambda}_{\text{imi}}$ and $\mathcal{\lambda}_{\text{rd}}$ are set to equal values, the training fails to converge. However, gradually increasing the weight $\mathcal{\lambda}_{\text{imi}}$ of $\mathcal{L}_{\text{imi}}$ leads to stable convergence and consistent performance gains. 

Further experiments demonstrate that the weight \(\lambda_{\text{sem}}\) of the auxiliary supervision loss \(\mathcal{L}_{\text{sem}}\) significantly influences model performance. When the auxiliary supervision is completely removed (i.e., \(\lambda_{\text{sem}}=0\)), the model performance degrades; conversely, when the weight of auxiliary supervision is excessively large, the performance gains diminish. These results indicate that while the auxiliary task plays an important role in model optimization, its influence should be subordinate to the primary planning task, necessitating a careful balance between the two.

\begin{table}[tb]
\caption{Ablation studies on different score functions \(\mathcal{S}(\cdot)\).
  }
  \label{tab:score_function}
  \centering
  \begin{tabular}{l|ccccc|c}
    \toprule
    
    \(\mathcal{S}(\cdot)\) & 
    {NC$\uparrow$} & 
    {DAC$\uparrow$} & 
    {EP$\uparrow$} & 
    {TTC$\uparrow$} & 
    {Comf.$\uparrow$} &  
    {EPDMS$\uparrow$}  \\
    \midrule
    
    DAC & 
    98.2 & 98.2 & 84.1 & 94.3 & 100 & 89.6 \\
    
    TTC & 
    98.4 & 97.9 & 83.4 & 95.0 & 100 & 89.6 \\

    PDMS & 
    98.4 & 97.9 & 83.7 & 95.0 & 100 & 89.7 \\
    
    \bottomrule
  \end{tabular}
\end{table}

\subsection{Ablation studies on score functions.}

We ablate various hard negative filtering scoring functions in \cref{tab:score_function}. Remarkably, using either DAC or TTC alone yields 89.6 PDMS, demonstrating that our method does not require complex multi-criteria evaluation. Furthermore, we observe that DAC-only filtering leads to models with elevated DAC performance, while TTC-only filtering produces models with superior TTC scores. This indicates that our framework supports targeted optimization of specific safety metrics.

\subsection{Experimental Results Details on HUGSIM Benchmerk.}

In \cref{tab:hugsim_more}, we present detailed experimental results of zero-shot evaluation on different datasets within the HUGSIM closed-loop benchmark (with models trained on the NAVSIM navtrain dataset). Our method achieves state-of-the-art (SOTA) performance across easy, medium, and extreme difficulty scenarios, which underscores both the generalizability and consistency of our approach.

\begin{table}[ht]
\centering

\caption{Zero-shot Performance on the HUGSIM Benchmark.}
\resizebox{\textwidth}{!}{
\begin{tabular}{l|c|cccccccc}
\toprule
\multirow{2}{*}{Method} 
& \multirow{2}{*}{Dataset}
& \multicolumn{2}{c}{Easy} 
& \multicolumn{2}{c}{Medium} 
& \multicolumn{2}{c}{Hard} 
& \multicolumn{2}{c}{Extreme} \\
\cmidrule{3-10}
 & & RC & HD-Score & RC & HD-Score & RC & HD-Score & RC & HD-Score \\
\midrule

\multirow{5}{*}{UniAD} & KITTI-360 & 16.6 & 4.7 & 11.7 & 1.9 & 12.2 & 1.7 & 9.4 & 0.6 \\
& Waymo & 78.4 & 66.4 & 54.7 & 41.9 & 59.0 & 40.1 & 32.6 & 17.1 \\
& nuScenes & 70.3 & 58.9 & 51.0 & 37.8 & 51.6 & 36.5 & 29.1 & 16.8 \\
& Pandaset & 69.1 & 64.7 & 47.3 & 36.6 & 38.9 & 30.8 & 32.9 & 22.6 \\
\cmidrule{2-10}
& \cellcolor{lightgray}Average & 
\cellcolor{lightgray}58.6 & 
\cellcolor{lightgray}48.7 & 
\cellcolor{lightgray}41.2 & 
\cellcolor{lightgray}29.5 & 
\cellcolor{lightgray}\textbf{40.4} & \cellcolor{lightgray}\textbf{27.3} & 
\cellcolor{lightgray}26.0 & 
\cellcolor{lightgray}14.3 \\
\midrule

\multirow{5}{*}{VAD} & KITTI-360 & 13.8 & 2.9 & 12.5 & 1.4 & 12.2 & 1.4 & 11.0 & 0.8 \\
& Waymo & 32.3 & 15.4 & 26.6 & 9.3 & 25.9 & 11.0 & 27.2 & 8.5 \\
& nuScenes & 54.0 & 34.8 & 34.0 & 13.2 & 38.3 & 20.4 & 31.1 & 15.8 \\
& Pandaset & 54.8 & 44.2 & 34.9 & 15.8 & 25.7 & 8.7 & 22.9 & 7.9 \\
\cmidrule{2-10}
& \cellcolor{lightgray}Average & 
\cellcolor{lightgray}38.7 & 
\cellcolor{lightgray}24.3 & 
\cellcolor{lightgray}27.0 & 
\cellcolor{lightgray}9.9 & 
\cellcolor{lightgray}25.5 & 
\cellcolor{lightgray}10.4 & 
\cellcolor{lightgray}23.0 & 
\cellcolor{lightgray}8.2 \\
\midrule

\multirow{5}{*}{MeanFuser} 
& KITTI-360 & 53.0 & 39.1 & 25.7 & 14.7 & 14.5 & 1.6 & 12.7 & 0.4 \\
& Waymo & 80.1 & 74.2 & 46.6 & 29.9 & 37.7 & 17.4 & 22.8 & 3.4 \\
& nuScenes & 84.1 & 72.1 & 53.5 & 32.5 & 65.6 & 51.8 & 41.4 & 21.1 \\
& Pandaset & 89.7 & 82.8 & 38.6 & 19.9 & 33.9 & 15.3 & 36.8 & 19.1 \\
\cmidrule{2-10}
& \cellcolor{lightgray}Average & 
\cellcolor{lightgray}76.7 & 
\cellcolor{lightgray}\textbf{67.1} & 
\cellcolor{lightgray}41.1 & 
\cellcolor{lightgray}24.3 & 
\cellcolor{lightgray}37.9 & 
\cellcolor{lightgray}21.5 & 
\cellcolor{lightgray}28.4 & 
\cellcolor{lightgray}11.0 \\
\midrule

\multirow{5}{*}{LTF} & KITTI-360 & 24.5 & 8.0 & 13.8 & 2.8 & 12.1 & 1.7 & 11.1 & 0.3 \\
& Waymo & 69.8 & 54.6 & 37.1 & 20.0 & 35.7 & 21.9 & 26.1 & 7.1 \\
& nuScenes & 84.7 & 61.6 & 54.2 & 30.7 & 49.8 & 22.6 & 35.7 & 17.1 \\
& Pandaset & 94.7 & 87.1 & 57.9 & 45.0 & 49.8 & 33.1 & 29.3 & 8.0 \\
\cmidrule{2-10}
& \cellcolor{lightgray}Average & 
\cellcolor{lightgray}68.4 & 
\cellcolor{lightgray}52.8 & 
\cellcolor{lightgray}40.7 & 
\cellcolor{lightgray}24.6 & 
\cellcolor{lightgray}36.9 & 
\cellcolor{lightgray}19.8 & 
\cellcolor{lightgray}25.5 & 
\cellcolor{lightgray}8.1 \\
\midrule

\multirow{5}{*}{LTFv7} & KITTI-360 & 55.7 & 43.3 & 21.7 & 19.2 & 15.1 & 9.3 & 12.0 & 2.4 \\
& Waymo & 79.5 & 70.7 & 41.0 & 25.3 & 42.8 & 24.1 & 23.0 & 13.2 \\
& nuScenes & 85.0 & 72.4 & 64.2 & 48.4 & 60.8 & 45.2 & 42.8 & 24.7 \\
& Pandaset & 87.2 & 76.2 & 48.2 & 32.3 & 33.3 & 26.7 & 40.7 & 24.4 \\
\cmidrule{2-10}
& \cellcolor{lightblue}Average & 
\cellcolor{lightblue}\textbf{76.8} & 
\cellcolor{lightblue}{65.5} & 
\cellcolor{lightblue}\textbf{43.0} & \cellcolor{lightblue}\textbf{31.4} & 
\cellcolor{lightblue}35.5 & 
\cellcolor{lightblue}26.3 & 
\cellcolor{lightblue}\textbf{29.6} & \cellcolor{lightblue}\textbf{16.2} \\

\bottomrule
\end{tabular}
}
\label{tab:hugsim_more}
\end{table}

\section{Negative Samples Analysis}

We present a detailed analysis of the characteristics of the generated hard negative samples. As shown in \cref{fig:dataset_analysis}(a), we count the number of samples with zero values for each sub-metric; collision-related indicators (NC and TTC) account for the vast majority. \Cref{fig:dataset_analysis}(b) illustrates the Euclidean distance between hard negative samples and expert demonstrations, exhibiting a clear long-tailed distribution where most hard negatives reside in close proximity to expert trajectories. \Cref{fig:dataset_analysis}(c) displays the correlation matrix among different safety metrics of the hard negative samples. Notably, NC and DAC show relatively low correlation, indicating two distinct unsafe patterns. Meanwhile, NC, DAC, and EP demonstrate the highest correlations with PDMS, suggesting that they dominate the failure modes.

\begin{figure}[tb]
  \centering
  \includegraphics[height=4.5cm]{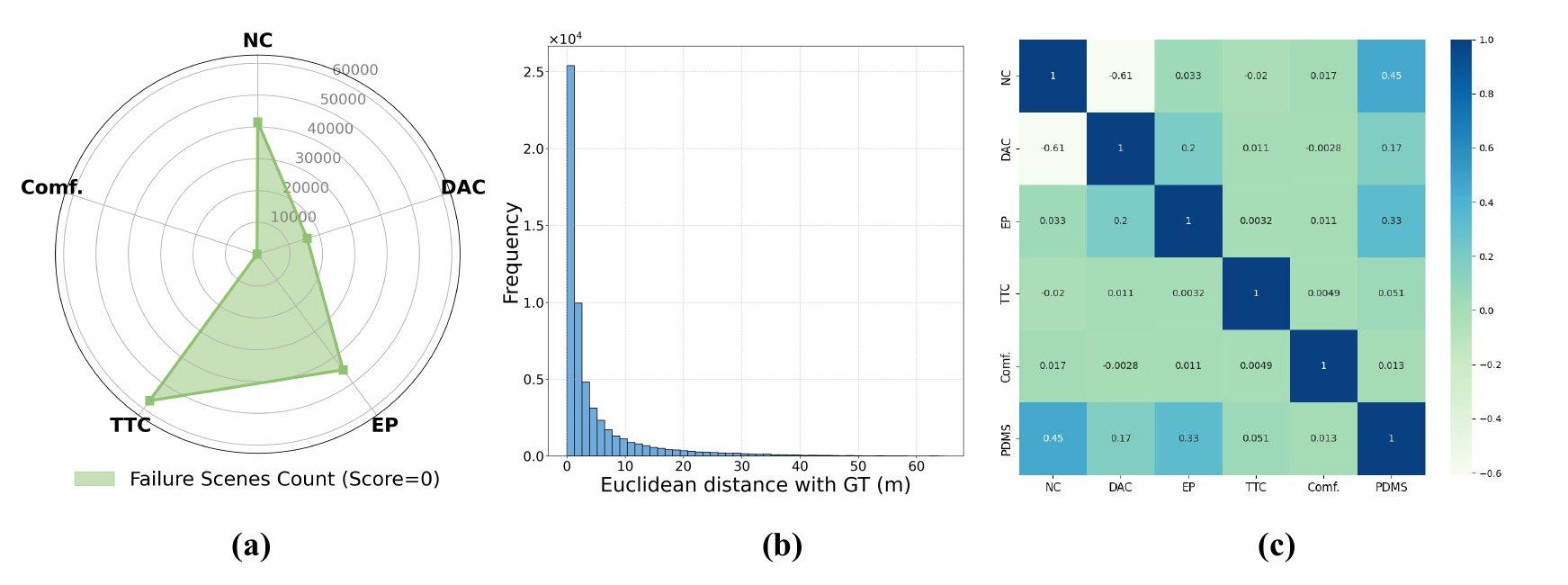}
  \caption{(a) Zero-value counts of hard negative samples. (b) Error distribution between hard negative samples and expert demonstrations. (c) Metric correlations of hard negative samples. 
  }
  \label{fig:dataset_analysis}
\end{figure}

\newpage

\section{Computing Resource Analysis}

\begin{wrapfigure}{l}{0.5\textwidth}
  \centering
  \includegraphics[width=\linewidth]{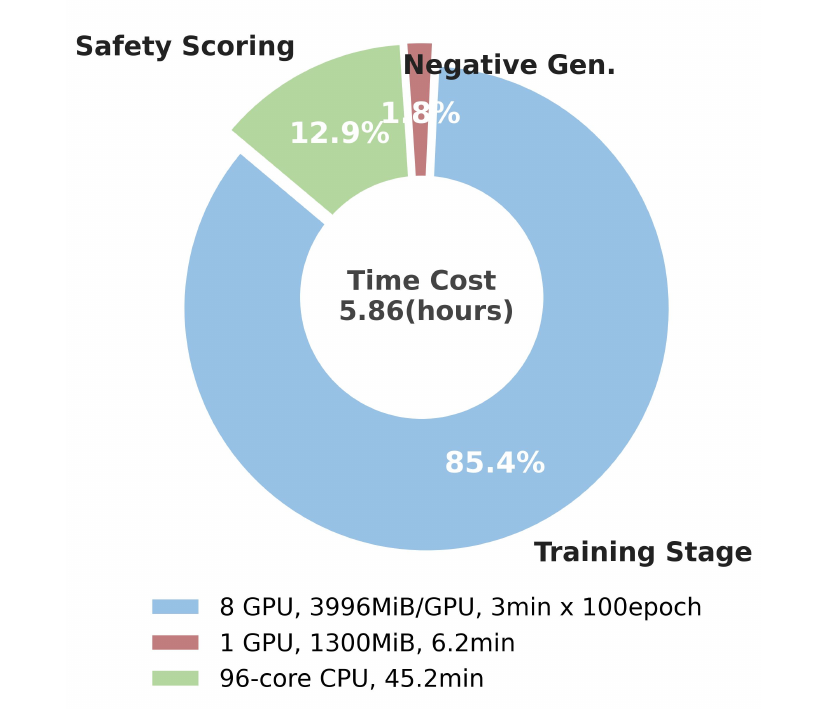}
  \caption{Time and resources.}
  \label{fig:resource_1}
  \vspace{-0.4cm}
\end{wrapfigure}

As shown in \cref{fig:resource_1}, we provide a detailed breakdown of the computational cost and resource consumption of the flow matching generator. The training phase spans 100 epochs on 8 GPUs, with a per-GPU batch size of 4 and the memory usage is 3.9GB. Each epoch takes approximately 3 minutes. During inference, we employ a single GPU with 1.3 GB memory usage, generating 64 trajectories per sample to build the negative sample pool. The trajectory safety assessment is executed on a 96-core CPU and requires 45.2 minutes. The total time cost amounts to approximately 5.86 hours.

\section{Qualitative Analysis}

\subsection{Visualization on HUGSIM Benchmerk.}

In \cref{fig:hugsim_scene_success_01} and \cref{fig:hugsim_scene_success_02}, we present two successful closed-loop scenarios where the ego vehicle perfectly navigates around an obscuring bus and a car ahead, and then returns to its current lane.

Additionally, in \cref{fig:hugsim_scene_faliure_01}, we show a failure scenario where an opposing vehicle intrudes into the ego vehicle's lane. Despite the ego vehicle's evasive maneuver, an unavoidable collision occurs because of the excessively aggressive behavior of the other agent. This indicates that the HUGSIM closed-loop testing environment presents significant adversarial difficulty.

\subsection{More Visualization on NAVSIM navtest Benchmerk.}

In \cref{fig:scene_go_straight}, \cref{fig:scene_turn_left}, and \cref{fig:scene_turn_right}, we respectively present the comparison of planning performance on the NAVSIM navtest dataset for straight-going, left-turn, and right-turn scenarios, demonstrating the improvement in driving safety achieved by our method.

\begin{figure}[tb]
  \centering
  \includegraphics[height=16.0cm]{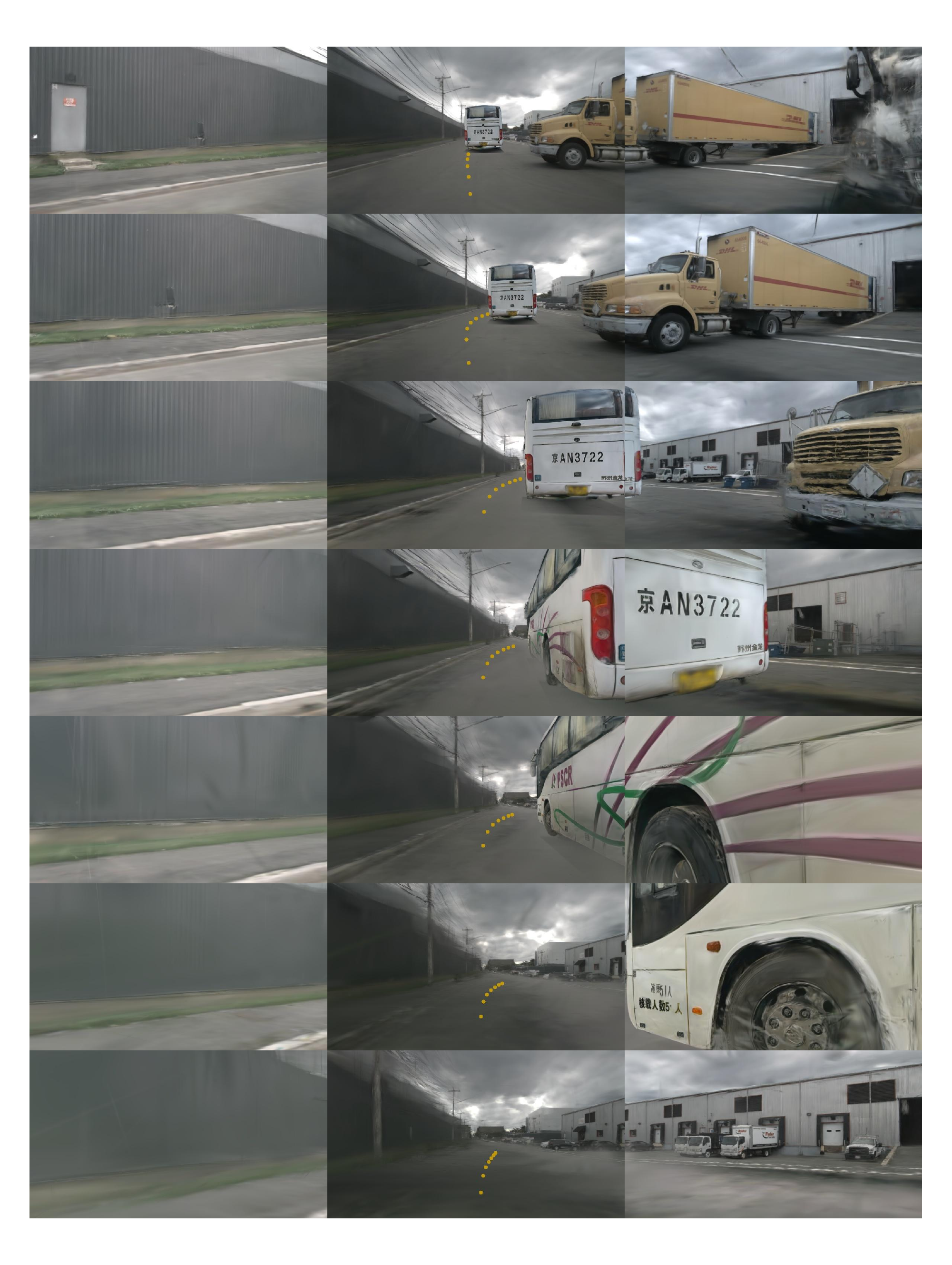}
  \caption{Case studies on the HuGSIM benchmark. The ego vehicle avoids a truck, overtakes a bus, and merges into the main road in a closed-loop environment.
  }
  \label{fig:hugsim_scene_success_01}
\end{figure}

\begin{figure}[tb]
  \centering
  \includegraphics[height=16.0cm]{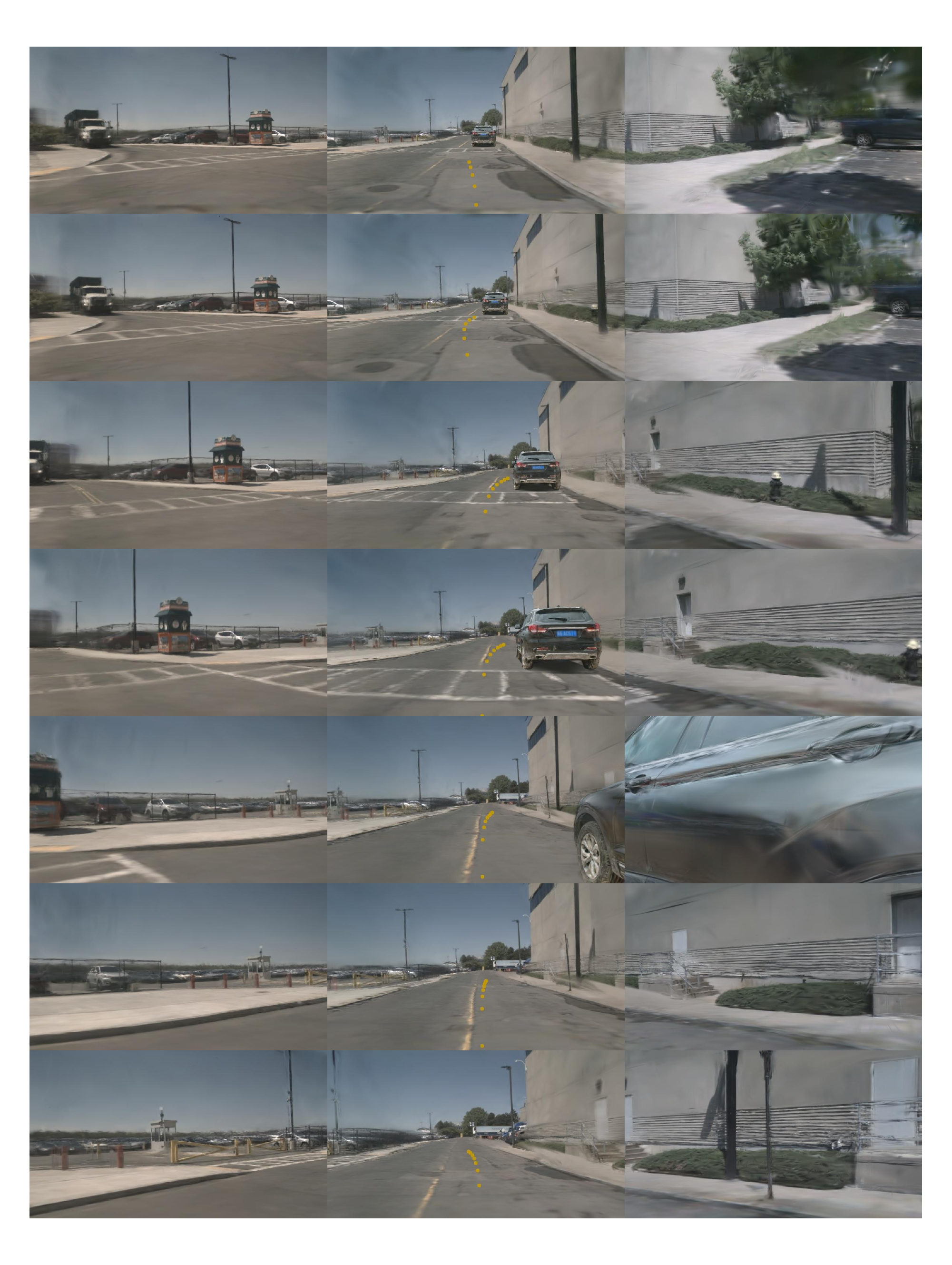}
  \caption{Case studies on the HUGSIM benchmark. The ego vehicle passes a pedestrian crossing, maneuvers around a stationary car on the roadside, and returns to its current lane in a closed-loop environment.
  }
  \label{fig:hugsim_scene_success_02}
\end{figure}

\begin{figure}[tb]
  \centering
  \includegraphics[height=16.0cm]{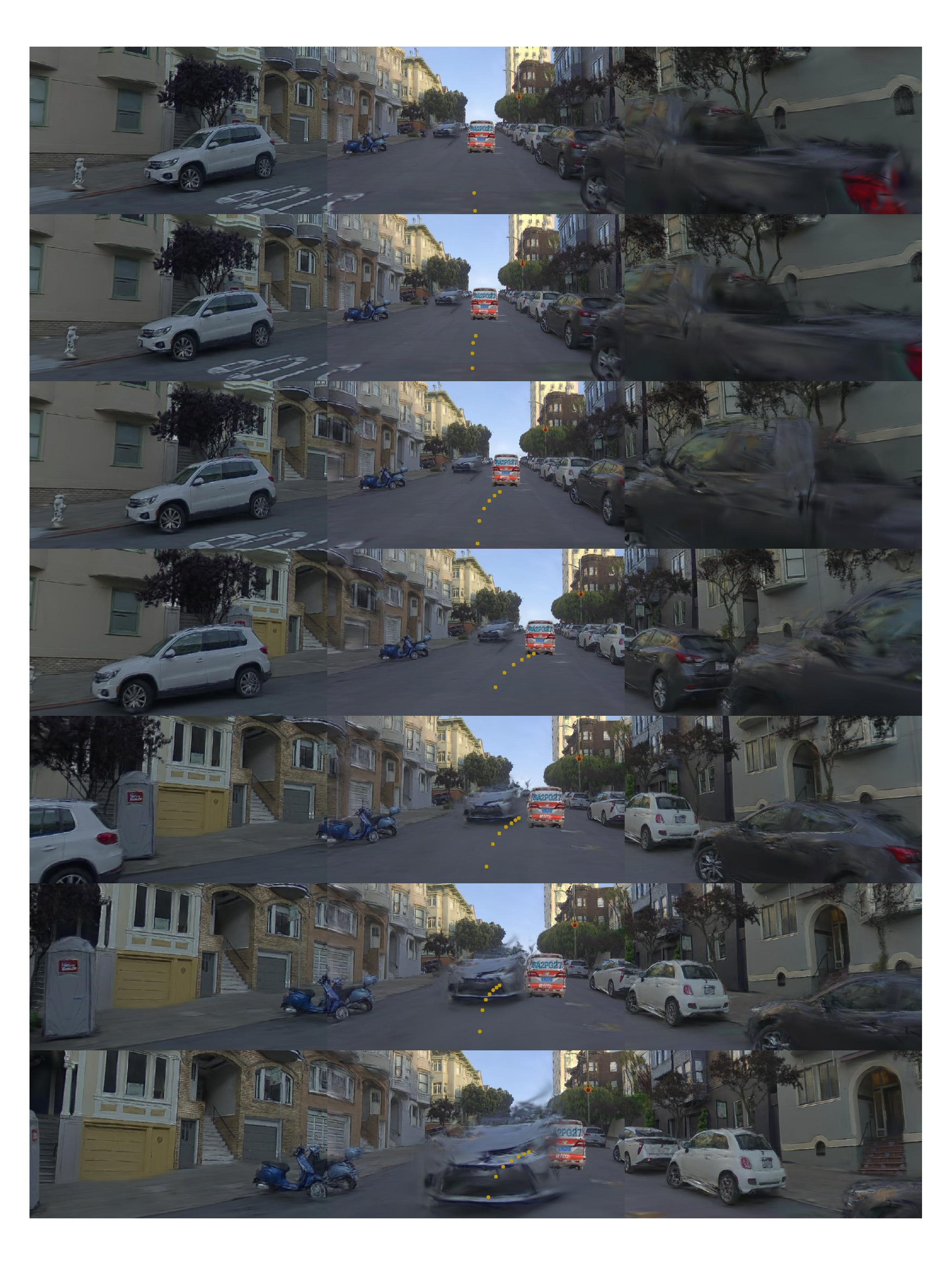}
  \caption{A failure case on the HUGSIM benchmark. In a meeting scenario, the oncoming vehicle swerves into the ego lane, and the ego vehicle fails to avoid the collision.
  }
  \label{fig:hugsim_scene_faliure_01}
\end{figure}

\begin{figure}[tb]
  \centering
  \includegraphics[height=14.0cm]{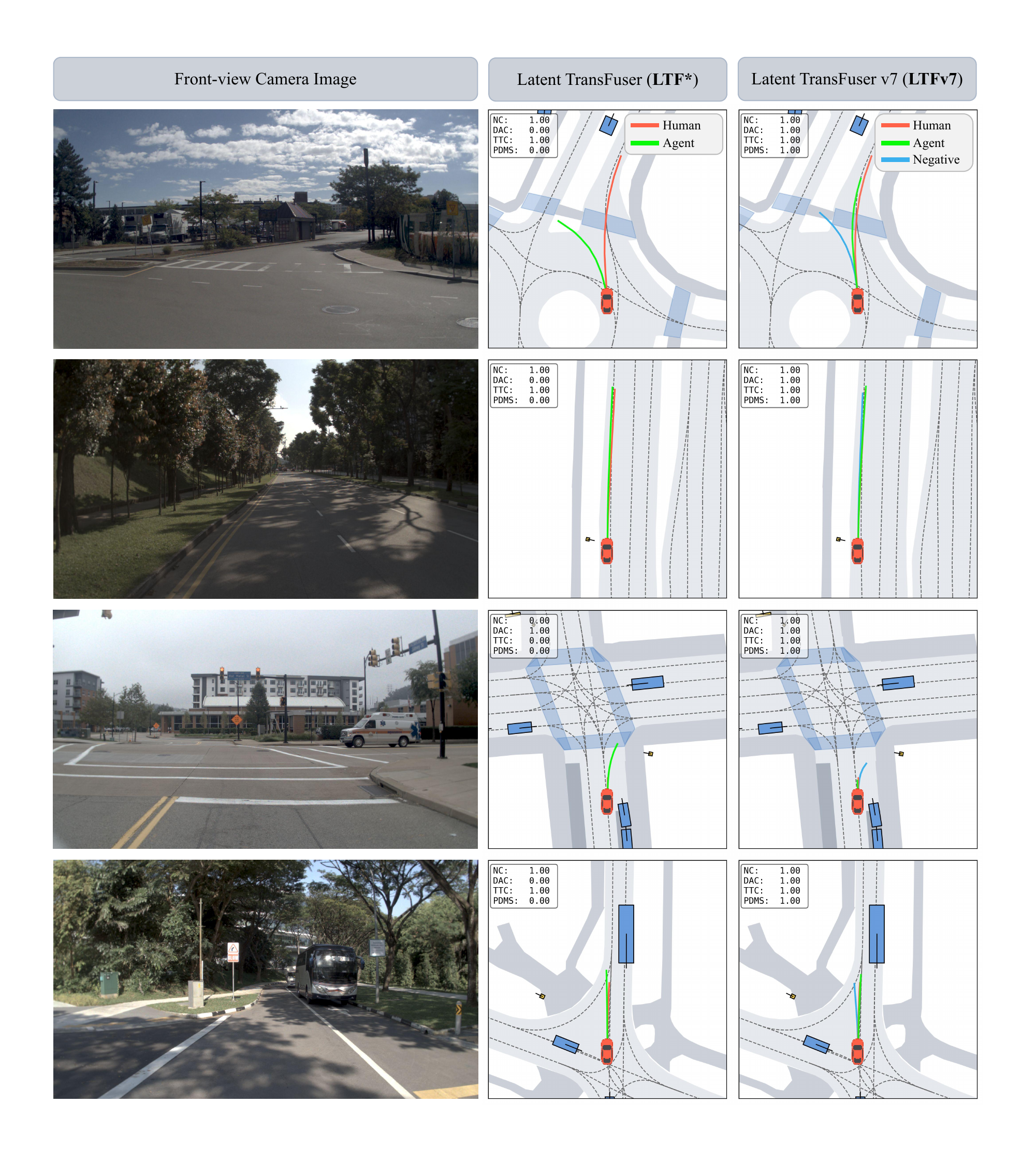}
  \caption{Comparison of planning performance in a straight driving scenario.
  }
  \label{fig:scene_go_straight}
\end{figure}

\begin{figure}[tb]
  \centering
  \includegraphics[height=14.0cm]{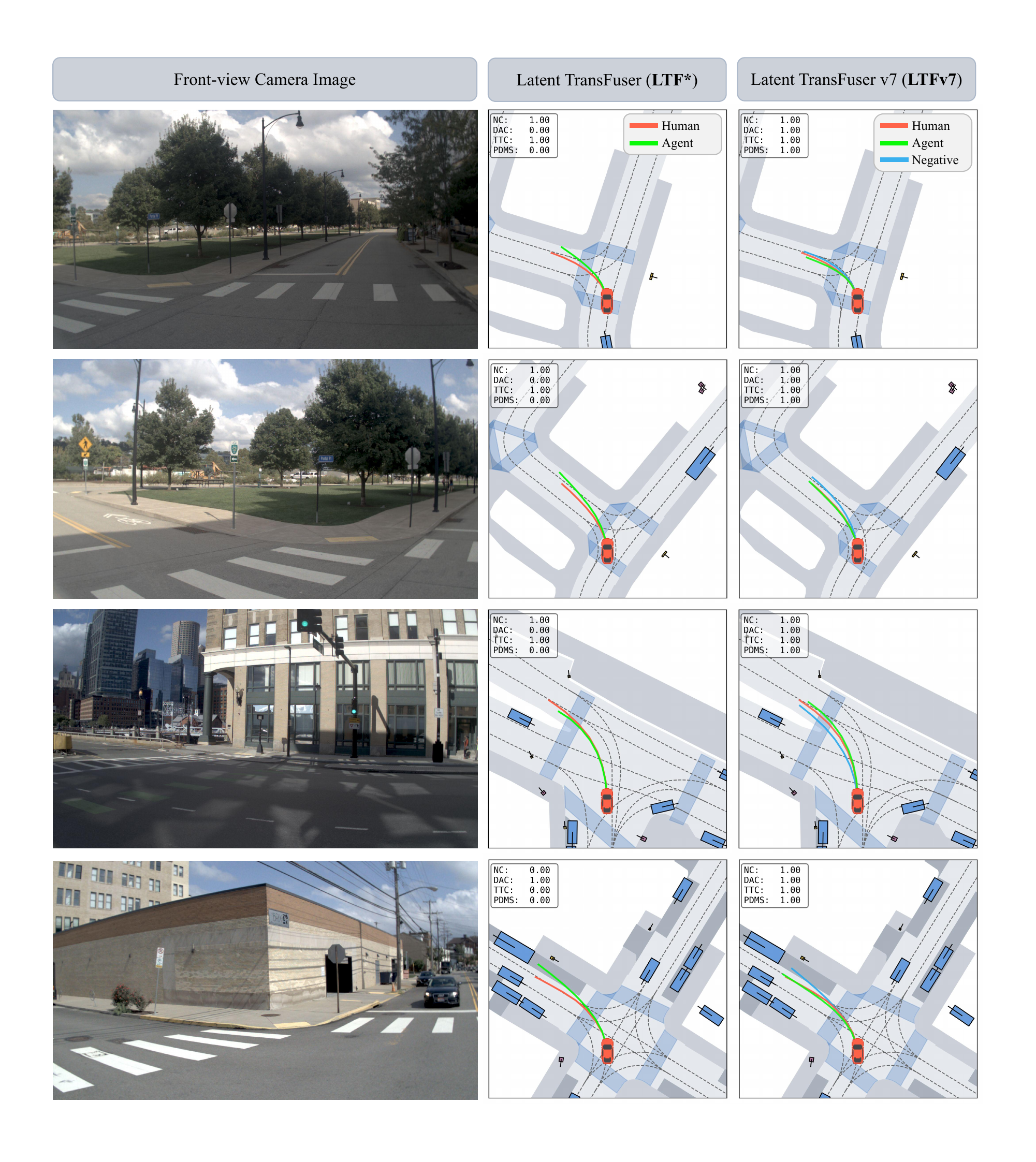}
  \caption{Comparison of planning performance in a left-turn scenario.
  }
  \label{fig:scene_turn_left}
\end{figure}

\begin{figure}[tb]
  \centering
  \includegraphics[height=14.0cm]{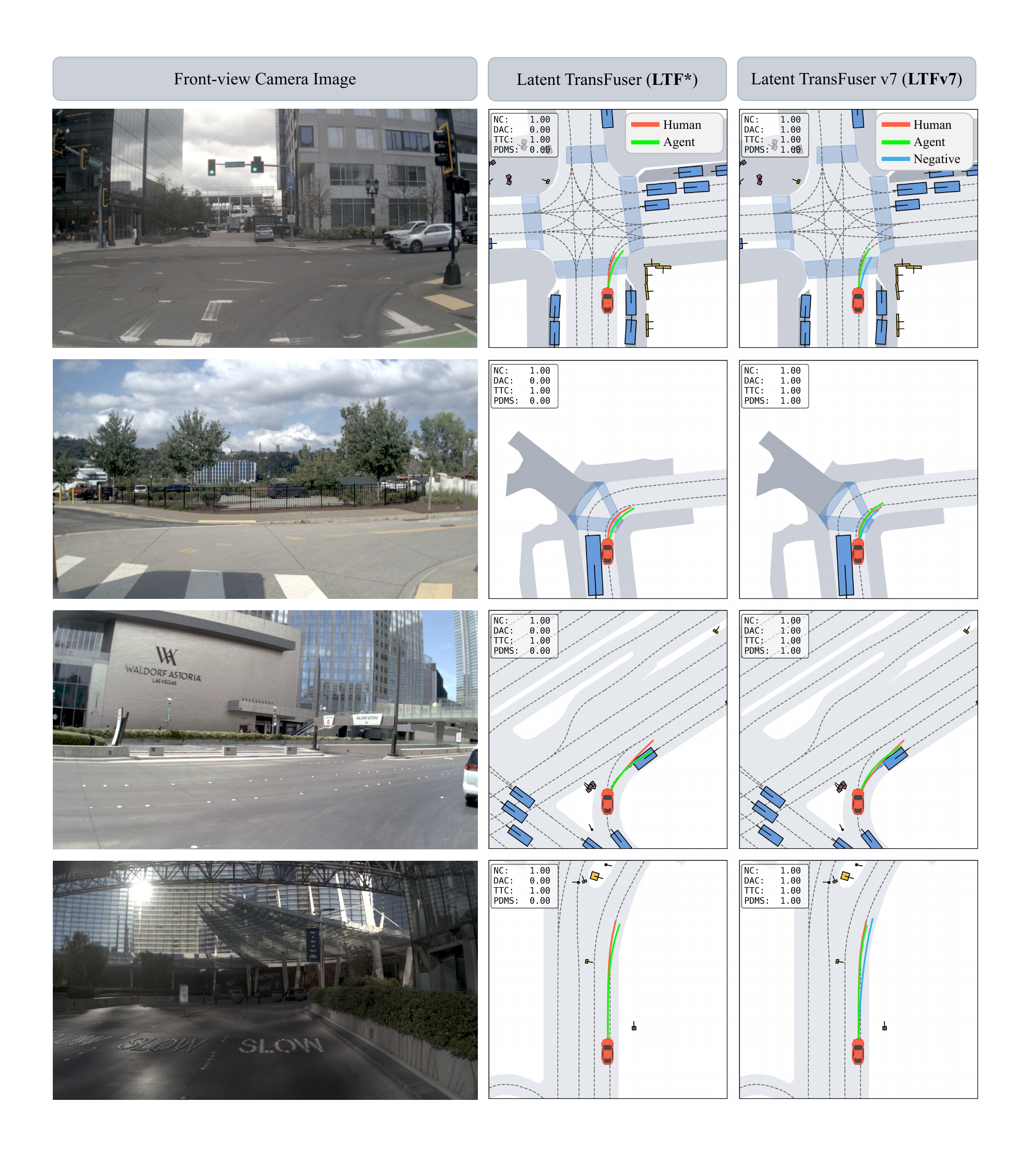}
  \caption{Comparison of planning performance in a right-turn scenario.
  }
  \label{fig:scene_turn_right}
\end{figure}

\end{document}